\documentclass{article} 
\usepackage{iclr2027_conference,times}

\usepackage{amsmath,amsfonts,bm}

\def\eqref#1{equation~\ref{#1}}

\def\1{\bm{1}}

\DeclareMathAlphabet{\mathsfit}{\encodingdefault}{\sfdefault}{m}{sl}
\SetMathAlphabet{\mathsfit}{bold}{\encodingdefault}{\sfdefault}{bx}{n}

\usepackage{hyperref}
\usepackage{url}

\usepackage{graphicx}
\usepackage[utf8]{inputenc} 
\usepackage[T1]{fontenc}    
\usepackage{hyperref}       
\usepackage{url}            
\usepackage{booktabs}       
\usepackage{amsfonts}       
\usepackage{amsmath}        
\usepackage{amssymb}        
\usepackage{amsthm}         
\usepackage{algorithm}      
\usepackage{algorithmic}    
\usepackage{listings}       
\usepackage{caption}        

\theoremstyle{remark}

\usepackage{nicefrac}       
\usepackage{microtype}      
\usepackage[table]{xcolor}  
\usepackage{afterpage}      
\usepackage{array}          
\usepackage{multirow}       
\usepackage{tablefootnote}  
\usepackage{tabularx}       
\usepackage{adjustbox}
\usepackage{subcaption}
\usepackage{makecell}
\usepackage[capitalize]{cleveref}
\usepackage{varwidth}
\usepackage{pgf}
\usepackage[rounding,point]{rccol}
\usepackage{wrapfig}
\usepackage{enumitem}

\newcommand{\T}{\textbf{\color{red}[T]}}

\newcommand{\Id}{\mathrm{Id}}

\definecolor{ourgreen}{RGB}{46, 204, 113}
\definecolor{ourgreenborder}{RGB}{39, 174, 96}
\definecolor{ourblue}{RGB}{52, 152, 219}
\definecolor{ourblueborder}{RGB}{41, 128, 185}
\definecolor{ourorange}{RGB}{230, 126, 34}
\definecolor{ourorangeborder}{RGB}{211, 84, 0}
\definecolor{ourred}{RGB}{231, 76, 60}
\definecolor{ourredborder}{RGB}{192, 57, 43}
\definecolor{ouryellow}{RGB}{241, 196, 15}
\definecolor{ouryellowborder}{RGB}{243, 156, 18}
\definecolor{ourpurple}{RGB}{155, 89, 182}
\definecolor{ourpurpleborder}{RGB}{142, 68, 173}
\definecolor{ourturquoise}{RGB}{26, 188, 156}
\definecolor{ourturquoiseborder}{RGB}{22, 160, 133}
\definecolor{ourturquoise}{RGB}{26, 188, 156}
\definecolor{ourturquoiseborder}{RGB}{22, 160, 133}
\definecolor{ourwhite}{RGB}{236, 240, 241}
\definecolor{ourwhiteborder}{RGB}{189, 195, 199}
\definecolor{ourgray}{RGB}{149, 165, 166}
\definecolor{ourgrayborder}{RGB}{127, 140, 141}

\definecolor{ourwhite2}{RGB}{246, 247, 248}

\definecolor{codecomment}{RGB}{120,120,120}
\definecolor{codekw}{RGB}{170,55,90}
\lstdefinestyle{pythonalg}{
  language=Python,
  basicstyle=\ttfamily\small,
  commentstyle=\color{codecomment}\itshape,
  keywordstyle=\color{codekw}\bfseries,
  numbers=none,
  showstringspaces=false,
  breaklines=true,
  columns=fullflexible,
  keepspaces=true,
  aboveskip=2pt,
  belowskip=2pt,
  xleftmargin=4pt,
  morekeywords={with,as,assert},
}

\definecolor{matplotlibblue}{HTML}{1f77b4}
\definecolor{matplotliborange}{HTML}{ff7f0e}
\definecolor{matplotlibgreen}{HTML}{2ca02c}

\makeatletter
\renewcommand{\cite}{%
  \PackageError{paper}{Do not use \string\cite. Use \string\citet\space or \string\citep\space (natbib) instead.}{}%
}
\makeatother

\newcommand{\rmd}{\mathrm{d}}

\newcommand{\tablestyle}[2]{\setlength{\tabcolsep}{#1}\renewcommand{\arraystretch}{#2}\centering\footnotesize}
\newcolumntype{x}[1]{>{\centering\arraybackslash}p{#1pt}}
\newcolumntype{y}[1]{>{\raggedright\arraybackslash}p{#1pt}}

\newcolumntype{H}{>{\setbox0=\hbox\bgroup}c<{\egroup}@{}}

\ExplSyntaxOn

\NewDocumentCommand{\roundtwofixed}{m}
  {
    \__my_ensure_two_decimals:e { \fp_eval:n { round(#1, 2) } }
  }

\cs_new:Npn \__my_ensure_two_decimals:n #1
  {
    \tl_if_in:nnTF {#1} {.}
      { \__my_pad_decimals:w #1 00 \q_stop }
      { #1.00 }
  }

\cs_generate_variant:Nn \__my_ensure_two_decimals:n { e }

\cs_new:Npn \__my_pad_decimals:w #1 . #2#3#4 \q_stop
  {
    #1.#2#3
  }

\ExplSyntaxOff

\newsavebox{\autosubtablebox}
\newcommand{\autosubtable}[3]{%
  \sbox{\autosubtablebox}{#3}%
  \vtop{%
    \hbox{%
      \captionsetup[subtable]{%
        width=\wd\autosubtablebox,
        margin=0pt%
      }%
      \subcaptionbox[t]{#1\label{#2}}[\wd\autosubtablebox][c]{%
        \usebox{\autosubtablebox}%
      }%
    }%
  }%
}

\newcommand{\boxhighlight}[2]{%
  \begingroup
  \rlap{\hspace{-\fboxsep}\colorbox{#1}{\phantom{#2}}}{#2}\vspace{-.3\fboxsep}%
  \endgroup
}

\title{Improved Distributional Diffusion Models}

\author{
  Tommaso Martorella$^{1,2}$ \; Alexandre Galashov$^{3,4}$ \; Felix Krause$^{1,2}$ \\
  \textbf{Stefan Andreas Baumann}$^{1,2}$ \, \textbf{Valentin De Bortoli}$^3$ \, \textbf{Arthur Gretton}$^{3,4}$ \, \textbf{Björn Ommer}$^{1,2}$\\
  $^1$CompVis @ LMU Munich \quad $^2$Munich Center for Machine Learning (MCML) \\
  $^3$Google DeepMind \quad $^4$Gatsby Unit @ UCL\\
}

\iclrfinalcopy 
\begin{document}

\maketitle
\begin{abstract}
  Distributional Diffusion Models (DDMs) replace the standard mean-prediction denoiser with a \emph{distributional} denoiser trained via a scoring rule objective, learning a stochastic approximation to $p(x_1 \mid x_t)$ rather than its conditional mean. However, scaling DDMs to modern image-generation settings faces two obstacles: (i) multi-particle training incurs overhead that scales with the number of particles, (ii) DDMs use globally fixed scoring rule hyperparameters, forcing a single trade-off across sampling budgets. We mitigate these limitations by deferring particle expansion to late transformer layers, and the hyperparameter trade-off by introducing time-dependent scoring rule schedules informed by the dynamical regimes of~\citet{Biroli2024}. Combined with a DiT-based latent setup, these changes make DDM training practical on class-conditional ImageNet-$256^2$, achieving 4.48 FID at 4 steps and 2.38 at 50 steps with DiT-XL/2, from a single model trained from scratch in one stage, without a teacher, self-distillation or JVPs. The result is a stochastic few-step generator whose FID does not degrade as the sampling budget grows from 4 to 50 NFE, and the same recipe transfers to text-to-image generation. Code and pre-trained models available at \url{https://github.com/CompVis/iDDM}.
\end{abstract}

\noindent

\begin{figure}[h]
  \centering
  \setlength{\tabcolsep}{0pt}
  \renewcommand{\arraystretch}{1}

  \begin{tabular}{@{}c@{\hspace{0.8em}}c@{}}
    \raisebox{14mm}{\rotatebox[origin=c]{90}{\textbf{4} steps}} &
    \includegraphics[
      width=0.86\linewidth,
      height=0.16\textheight,
      keepaspectratio
    ]{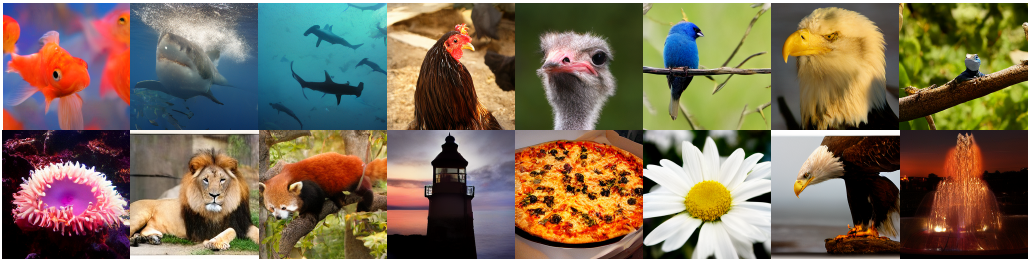}\\[1.0ex]

    \raisebox{14mm}{\rotatebox[origin=c]{90}{\textbf{50} steps}} &
    \includegraphics[
      width=0.86\linewidth,
      height=0.16\textheight,
      keepaspectratio
    ]{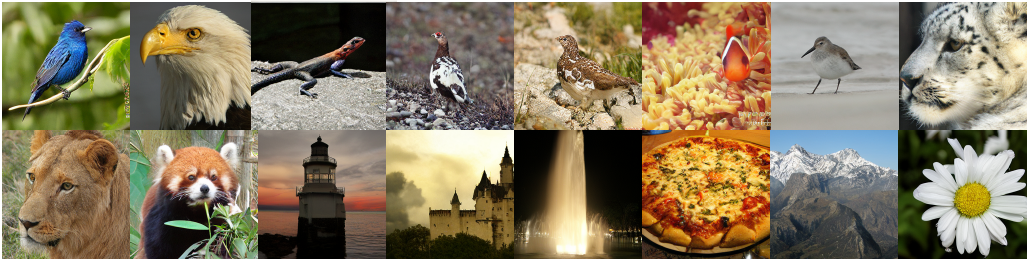}
  \end{tabular}

  \caption{
    \textbf{Improved Distributional Diffusion Models (iDDM)} achieve competitive sample quality across 4–50 steps with a single model and an FID on ImageNet-256$^2$ of 4.48 and 2.38 while offering a \emph{single-stage} training without distillation, CFG during training or self-bootstrapping.
  }
  \label{fig:teaser}
\end{figure}
\vskip 0.06in

\section{Introduction}
\label{sec:intro}

Diffusion models~\citep{pmlr-v37-sohl-dickstein15, NEURIPS2020_4c5bcfec, song2021scorebased} and flow-based variants~\citep{lipman2023flow, liu2023flow, albergo2025stochastic} have established themselves as a leading paradigm in generative modeling and found many applications in high-quality synthesis of images~\citep{rombachHighResCVPR,saharia2022photorealistic}, video~\citep{ho2022video,polyak2025moviegencastmedia}, audio~\citep{kong2021diffwave, pmlr-v202-liu23f} and proteins \citep{watson2023novo,abramson2024accurate}.
Adopting flow-matching notation ($p_0$ : noise, $p_1$: data), samples are generated by numerically integrating an Ordinary Differential Equation from noise to data. High-fidelity approximation of this dynamics typically requires a large number of function evaluations (NFE).
This becomes especially challenging in the few-step regime, where the conditional law $p(x_1|x_t)$ can exhibit substantial variance at coarse discretizations. In this regime, trained denoisers which model the conditional mean $\mathbb{E}[x_1 \mid x_t]$ provide an insufficient summary of the posterior distribution.

Most existing approaches reduce this cost by improving the numerical solver~\citep{karras2022elucidating, lu2022dpmsolver, Lu_2025, zheng2023dpmsolverv}, distilling a multi-step model~\citep{luhman2021knowledge, salimans2022progressive, sauer_distillation, Yin_2024_CVPR, salimans2024multistep}, or directly training a few-step generator, often with consistency- or flow-map-based self-distillation objectives~\citep{pmlr-v202-song23a, frans2025one, boffi2025how, dieleman2026flowmaps}. Distributional Diffusion Models (DDMs)~\citep{debortoli2025distributional} pursue a complementary direction: they replace conditional-mean regression with a \emph{stochastic} denoiser $\hat x_\theta(t, x_t, \xi)$ that, for each auxiliary noise $\xi \sim \mathcal{N}(0,\Id)$, returns a sample from a learned approximation to $p(x_1 \mid x_t)$. The model is trained with a generalized energy score parameterized by an exponent $\beta \in (0,2]$ and a trade-off weight $\lambda \in [0,1]$, with a confinement term $\|x_1-\hat{x}_{\theta}(x_t,t,\xi)\|^{\beta}_{2}$ which encourages the fidelity to the target, while an interaction term, weighted by $\lambda$, penalizes collapse of the sample population to a single prediction. In the limiting case, setting $\lambda=0,\beta=2$ recovers the standard MSE diffusion loss. DDMs thus offer a non-adversarial alternative to modeling the full conditional distribution, not only its mean.

Despite their promise, DDMs remain impractical at modern model scale. We therefore ask: \emph{what prevents distributional denoising from scaling, and what is needed to make it practical?} We identify two bottlenecks, both of which must be removed for DDMs to scale. \textbf{First}, the DDM objective draws $m$ noise samples $\xi$ per training example, with each particle requiring a separate denoiser evaluation, resulting in an $O(m)$ computational overhead. Yet most of this computation extracts the same features from $x_t$. \textbf{Second}, the original DDM uses fixed scoring rule hyperparameters $(\lambda,\beta)$ across all noise levels, imposing the same fidelity–diversity trade-off throughout the diffusion trajectory, including near $t\approx 1$, where $p(x_1\mid x_t)$ becomes concentrated, so that no single setting serves both coarse and fine sampling budgets (as \citet{debortoli2025distributional} showed both in the Gaussian case and on ImageNet).
We remove both bottlenecks with two orthogonal modifications:
\begin{itemize}
    \item \textbf{Reduced particle cost:} we process $x_t$ once through most of the transformer and expand into $m$ particles only in the final few layers, reducing the marginal cost of additional particles to a fraction of a full forward pass.
    \item \textbf{Adaptive objective:} we replace fixed $(\lambda,\beta)$ with time-dependent schedules $(\lambda(t),\beta(t))$ that tie the scoring rule to how concentrated $p(x_1\mid x_t)$ is at each noise level. This lets the model emphasize diversity when $p(x_1\mid x_t)$ is broad and more concentrated predictions when it is narrow following the work of \citet{Biroli2024}. 
\end{itemize}  

Removing both bottlenecks makes distributional denoising practical with large transformer backbones while retaining the goal of modeling $p(x_1\mid x_t)$ rather than only its conditional mean. The result is a stochastic few-step generator, trained from scratch in a single stage, whose FID is non-increasing from 4 to 50 steps, so one checkpoint serves every budget in that range. We demonstrate results on class-conditional ImageNet-$256^2$ with a DiT backbone~\citep{peebles2023dit} and show that the same approach transfers to text-to-image (T2I), where iDDM nearly halves the 4-step MS-COCO FID of a matched flow-matching baseline (41.05 vs.\ 78.20).

\section{Background}
\label{sec:background}

\subsection{Flow Matching}
\label{sec:flow_matching}

Flow matching is a specific instantiation of diffusion models~\citep{lipman2023flow, albergo2025stochastic, liu2023flow} where we learn a time-dependent velocity field $v_\theta(t, x_t)$ that transports samples from $p_0$ (noise) to $p_1$ (data) distributions. Given a data sample $x_1 \sim p_\text{data}$ and noise $x_0 \sim \mathcal{N}(0, \Id)$, we define the interpolant
\begin{equation}\label{eq:interpolant}
    \textstyle
    x_t = (1 - t)\, x_0 + t\, x_1, \quad t \in [0, 1]\,,
\end{equation}
The model is trained by regressing on the conditional velocity field $u_t(x_t | x_0, x_1) = x_1 - x_0$ via
\begin{equation}\label{eq:fm_loss}
    \textstyle
    \mathcal{L}_\text{FM}(\theta) = \mathbb{E}_{t, x_0, x_1}\left[\|\hat{v}_\theta(t, x_t) - u_t(x_t | x_0, x_1)\|^2\right]\,.
\end{equation}
The minimizer of this objective is the conditional mean velocity, i.e. $\hat{v}_{\theta^\star}(t, x_t)\approx \mathbb{E}[x_1-x_0\mid x_t]$.
At inference, samples are generated by integrating the learned velocity field from $t{=}0$ to $t{=}1$ using a numerical ODE solver (\emph{e.g.}, Euler). For $K$ steps with uniform spacing $\Delta t = 1/K$, starting from $x_0 \sim \mathcal{N}(0,\Id)$, it yields the following iterative procedure.
\begin{equation}\label{eq:euler}
    \textstyle
    x_{t + \Delta t} = x_t + \Delta t \cdot \hat{v}_{\theta^\star}(t, x_t)\,.
\end{equation}

\subsection{Distributional Diffusion Models}
\label{sec:ddm_background}

Standard diffusion models learn $\hat{x}_\theta(t, x_t) \approx \mathbb{E}[x_1 | x_t]$, the conditional mean of the clean data, connected to the velocity field via $\hat{x}_\theta(t, x_t)  = x_t + (1-t) \hat{v}_\theta(t, x_t)$. For sufficiently fine time discretization (large $K$), the Euler iterates~\eqref{eq:euler} with this learned model approximate the target distribution $p_{\text{data}}$; see e.g. \citep{song2021scorebased}. However, for coarse discretization (small $K$), this guarantee no longer holds.

Distributional Diffusion Models (DDMs)~\citep{debortoli2025distributional} address this by learning a generative model $\hat{x}_\theta(t, x_t, \xi)$ that takes an additional noise input $\xi \sim \mathcal{N}(0,\Id)$ and aims to approximate $p(x_1 | x_t)$. It is trained  by maximizing a \emph{generalized energy score}~\citep{debortoli2025distributional,Gneiting01032007,NIPS2016_c0e190d8} parameterized by $\beta \in (0, 2]$ and $\lambda \in [0,1]$:
\begin{equation}
\label{eq:kernel_score}
   \textstyle S_{\lambda, \beta}(p, y) = -\mathbb{E}_p\!\left[ \|X -  y \|^{\beta}_{2}\right]\, + \frac{\lambda}{2}\, \mathbb{E}_{p \otimes p}\!\left[\| X - X' \|^{\beta}_{2}\right],
\end{equation}
where $p$ is the learned distribution, $y \sim q$ is a sample from a target distribution. In \eqref{eq:kernel_score} $\lambda \in [0,1]$ controls the relative weight of the interaction term. When $\beta \in (0,2)$ and $\lambda =1$, $S_{\lambda, \beta}$ is strictly proper~\citep{Gneiting01032007} with the unique maximizer equal to the target distribution, while $\beta=2, \lambda=0$ recovers the standard MSE regression objective, targeting only the conditional mean.

The DDM training loss integrates this score over diffusion time, yielding the following objective
\begin{equation}\label{eq:ddm_loss}
  \textstyle  \mathcal{L}_\text{DDM}(\theta) = -\int_0^1 w(t)\, \mathbb{E}_{p(t)} \left[ \mathbb{E}_{p(x_0, x_1, x_t)}\!\left[S_{\lambda, \beta}\!\left(p^\theta(\cdot \mid x_t),\, x_1\right)\right] \right] \rmd t\,,
\end{equation}
where $w(t)$ is a loss weighting function and $p(t)$ is a time distribution (typically uniform). In practice, $p^\theta(\cdot \mid x_t)$ is represented by $\{\hat{x}_\theta(t, x_t, \xi_j)\}_{j=1}^m$ with $m$ samples $\xi_j \overset{\mathrm{i.i.d.}}{\sim} \mathcal{N}(0,\Id)$. The loss~\eqref{eq:ddm_loss} can equivalently be expressed in terms of a velocity field $\hat{v}_\theta(t, x_t,\xi)$ using the relationship
\begin{equation}
    \textstyle
    \label{eq:stochastic_velocity}
    \hat{x}_\theta(t, x_t, \xi) = x_t + (1-t) \hat{v}_\theta(t, x_t,\xi)
\end{equation}
At inference, a single $\xi \sim \mathcal{N}(0,\Id)$ is drawn per denoising step. The stochastic prediction $\hat{x}_\theta(t, x_t, \xi)$ induces a stochastic velocity $\hat{v}_\theta(t, x_t,\xi)$ via~\eqref{eq:stochastic_velocity}, which is then used in an Euler update (Eq.~\ref{eq:euler}). 

\subsection{Dynamical Regimes of the Reverse Process}
\label{sec:dynamical_regimes_background}

\citet{Biroli2024} study the Ornstein--Uhlenbeck forward diffusion process\footnote{For consistency with \citet{Biroli2024}, we introduce the dynamical regimes using the diffusion model convention but will later use them using the flow matching convention of Section \ref{sec:flow_matching}.}
\[
    \textstyle
    \rmd x_\tau = -x_\tau\,\rmd\tau + \sqrt{2}\,\rmd W_\tau,
    \qquad x_{0} \sim p_{\mathrm{data}},
\]
for which the noised distribution is obtained by convolution with a Gaussian of variance
$\Delta_\tau := 1-\mathrm{e}^{-2\tau}$.
Under the exact empirical score assumption, they analyze the corresponding reverse dynamics in the large-dimensional regime and show that it passes through three regimes separated by two characteristic times. The first transition is the \emph{speciation time} $\tau_s$, defined by $\Lambda \mathrm{e}^{-2\tau_s} = 1$.
where $\Lambda$ is the top eigenvalue of the data covariance matrix
$C_0 = \mathbb{E}[X_0 X_0^\top]$ for  $X_0 \sim p_{\text{data}}$ (assuming that it is centered).
At this point, coarse structure in the data becomes distinguishable from noise. The second transition is the \emph{collapse time} $\tau_c$, defined by $H(\tau_c)=H^{\mathrm{sep}}(\tau_c)$
where $H(\tau)=-\frac{1}{d}\int \rmd x\, p_\tau(x)\log p_\tau(x)$ is the per-dimension entropy of $p_\tau$, and
$H^{\mathrm{sep}}(\tau)=\frac{\log n}{d}+\frac{1}{2}+\frac{1}{2}\log(2\pi\Delta_\tau)$ is the entropy of $n$ well-separated Gaussians with variance $\Delta_\tau$ ($n$ : dataset size).

These two transition times define three regimes: before speciation, the reverse process remains essentially undifferentiated; between $\tau_s$ and $\tau_c$, the dynamics captures coarse structure without collapsing onto a training example; after collapse, a trajectory is attracted toward a particular training datum.  The mapping to our parameterization and the associated schedule construction are in Sections~\ref{sec:time_dependent_scoring_rule_schedules} and Appendix~\ref{app:dynamical_regimes}.
\section{Method}
\label{sec:method}
\vspace{-0.1in}

We build on DDMs in two orthogonal directions. Section~\ref{sec:efficient_training} reduces the cost of multi-particle training through deferred \emph{population expansion} and the associated $\xi$ conditioning. Section~\ref{sec:time_dependent_scoring_rule_schedules} modifies the training objective through time-dependent scoring rule schedules.

\subsection{Efficient Multi-particle Training and $\xi$ Conditioning}
\label{sec:efficient_training}

\begin{figure}[t]
  \centering
  \includegraphics[scale=1.35]{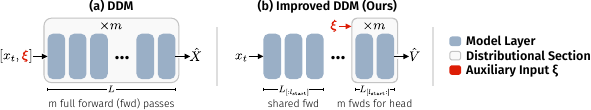}
  \caption{\textbf{Deferred Population Expansion.} \textbf{(a)} Standard DDM replicates the input $m$ times and processes all particles with the full model ($m{\times}$ cost), concatenating $\xi$ at the input. \textbf{(b)}~Our~approach processes the shared representation through layers $0, \ldots, \ell_\text{start}{-}1$ once, then replicates and injects $\xi$ conditioning at layer $\ell_\text{start}$. Only the remaining $L - \ell_\text{start}$ layers pay the $m{\times}$ cost.}
  \label{fig:late_expansion}
  \vspace{-0.1in}
\end{figure}

\paragraph{Multi-Particle Training.} The DDM objective (\cref{eq:ddm_loss}) requires, for each training example, a population of $m$ predictions $\{\hat v_\theta(t,x_t,\xi_j)\}_{j=1}^m$ (Eq.~\ref{eq:stochastic_velocity}) obtained from the same $(x_t,t)$ with independent auxiliary noises $\xi_1,\dots,\xi_m$. We call this replication \emph{population expansion}. Naïvely, expansion happens at the network input, so the transformer reprocesses the same $x_t$ $m$ times (Fig.~\ref{fig:late_expansion}), although particle-specific variation enters only through $\xi_j$. We instead \emph{defer} expansion: for a minibatch of size $B$, layers $0,\dots,\ell_{\text{start}}-1$ run once at batch size $B$; the hidden state is then replicated, $\xi$ is injected, and layers $\ell_{\text{start}},\dots,L-1$ run at batch size $B\times m$ (Fig.~\ref{tab:xi_conditioning_hparam_ablation}). Unlike shared-trunk architectures with independently parameterized heads, all particles pass through the same remaining blocks and differ only in the injected noise. This reduces training-step compute to $\big(\ell_{\text{start}}+(L-\ell_{\text{start}})m\big)/(Lm)$ of naïve DDM, i.e.\ $\approx36\%$ for DiT-XL ($L{=}28$, $\ell_{\text{start}}{=}24$, $m{=}4$); we ablate $\ell_{\text{start}}$ in~\cref{sec:ablation}.

\begin{figure}
  \centering
  \includegraphics[scale=1.35]{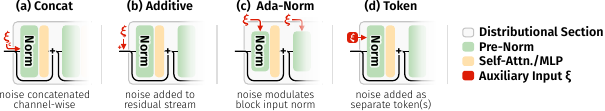}
    \caption{
    \textbf{\(\xi\)-conditioning mechanisms.}
    After expansion, \(\xi\) is injected by \textbf{(a)} channel-wise concatenation, which appends \(\xi\)-features to the residual stream, \textbf{(b)} residual addition, which adds projected \(\xi\)-features to hidden states, \textbf{(c)} AdaNorm modulation, which conditions the normalization path, \textbf{(d)} or register tokens, which expose \(\xi\) through additional tokens.
    }
  \label{fig:xi_conditioning}
  \vspace{-0.2in}
\end{figure}

\paragraph{$\xi$ Conditioning.} After expansion, each particle's noise $\xi_j$ must be injected into the replicated hidden state. We compare four mechanisms (Fig.~\ref{fig:xi_conditioning}): channel-wise concatenation, residual addition, adaptive normalization, and register tokens. Concatenating $\xi_j$ widens the residual stream from $d$ to $d+d_{\text{cat}}$ channels, so we test three variants: (i) widening all late-layer attention and MLP projections accordingly, (ii) widening only the residual stream while keeping attention and MLP at their original inner width (\emph{fixed inner width}), or (iii) replacing $d_{\text{cat}}$ existing channels with $\xi_j$ so the width stays $d$. We adopt fixed-inner-width concatenation, as it gives the lowest 4-step FID at roughly the baseline parameter count, whereas adaptive normalization and register tokens break down at few steps (Tab.~\ref{tab:xi_conditioning_hparam_ablation}). Concretely, $h_j=[h,\xi_j]\in\mathbb{R}^{N\times(d+d_{\text{cat}})}$ for $N$ tokens is carried through all remaining blocks, and the velocity is read from the first $d$ channels. A learned $t$-dependent gate (Eq.~\ref{eq:adaptive_ksi_gating}, App.~\ref{app:xi_ablation}) further lets the model scale $\xi_j$ per noise level. All choices are ablated in Section~\ref{sec:ablation}.

\subsection{Time-dependent Scoring Rule Schedules}
\label{sec:time_dependent_scoring_rule_schedules}

\citet{debortoli2025distributional} found that the best $(\lambda,\beta)$ (Eq.~\ref{eq:ddm_loss}) depends on the sampling budget: models trained with $\lambda\approx1,\beta\approx1$ sample best with few steps, whereas $\lambda\approx0,\beta\approx2$ is best with many steps. A key insight from DDM is that $\lambda <1$ causes the model to underestimate the variance of $p(x_1|x_t)$, biasing learning toward the conditional mean. When the posterior is already concentrated (e.g. $t \approx 1$), this bias is beneficial; when it is broad, it is harmful. This motivates making $(\lambda,\beta)$ depend on $t$.

We propose time-dependent schedules $(\lambda(t),\beta(t))$ to allocate the model's finite capacity by linking the training objective to the complexity of $p(x_1 \mid x_t)$ at each phase of the diffusion process, using the dynamical regimes of~\citet{Biroli2024} to locate where the objective should transition from distributional toward regression-like behavior. App.~\ref{app:on_lambda_beta} further motivates this orientation through a posterior-covariance bias criterion, which determines the direction of the schedule not its exact shape.

\subsubsection{Understanding the Posterior Evolution Through Dynamical Regimes}
\label{sec:dynamical_regimes}

As described in Section~\ref{sec:dynamical_regimes_background}, the dynamical regimes framework of \citet{Biroli2024} identifies three phases of the reverse dynamics separated by two characteristic transition times. We use these transition times as anchors for schedule design, linking the dynamical picture to the evolution of the conditional law $p(x_1\mid x_t)$. In \citet{Biroli2024}, we have that $x_\tau = \exp[-\tau] x_{\mathrm{data}} + (1 - \exp[-2\tau])^{1/2} z$ whereas in \eqref{eq:interpolant} we have $x_t = (1-t) z + t x_{\mathrm{data}}$. In order to transfer the learning of \citet{Biroli2024} to our setting we perform the time change of variable $t \mapsto \phi(t)$ such that $\rho(t) = \frac{\exp[-2 \phi(t)]}{1 - \exp[-2\phi(t)]} = \frac{t^2}{(1-t)^2}$, 
thereby effectively mapping the Signal-to-Noise Ratio (SNR) of Flow Matching to the one of \citet{Biroli2024}. 

For schedule design, we therefore use the following regime picture:

\textbf{Regimes I-II ($\rho<\rho_c$): Prior and Manifold Localization.} Below speciation ($\rho<\rho_s$), $p(x_1\mid x_t)$ is close to the data prior; between speciation and collapse it localizes onto a lower-dimensional manifold as a complex, multimodal distribution. \\
\boxhighlight{gray!10}{\textbf{Scheduling Implication (Regime I and II):} Favor a more distributional objective.} \\
\textbf{Regime III ($\rho\ge\rho_c$): Collapse.} After collapse, the posterior is mode-dominated but not necessarily concentrated. We refine this regime with a geometric separation threshold $\rho_{\mathrm{sep}}(\varepsilon)=4\chi^2_{d,1-\varepsilon}/d_{\mathrm{min}}^2$, where $d_{\mathrm{min}}$ is the minimum pairwise distance between training points and $\chi^2_{d,1-\varepsilon}$ the $(1-\varepsilon)$-quantile of a $\chi^2_d$ variable: at this SNR, the $(1-\varepsilon)$-quantile noise ball around each noised training point fits within half the distance to its nearest neighbor (derivation in App.~\ref{app:regime_thresholds}). This splits Regime III into an ambiguous window IIIa ($\rho_c\le\rho<\rho_{\mathrm{sep}}$) and an unambiguous data-end regime IIIb ($\rho\ge\rho_{\mathrm{sep}}$). \\
\boxhighlight{gray!10}{\textbf{Scheduling Implication (Regime IIIa):} Maintain spread while tapering to a sharper objective.}\\
\boxhighlight{gray!10}{\textbf{Scheduling Implication (Regime IIIb):} Regression-dominated objective.}

\label{sec:schedules}

We instantiate time-dependent schedules for the generalized energy score~\eqref{eq:kernel_score}, varying both the interaction weight $\lambda\in[0,1]$ and the norm exponent $\beta\in(0,2]$ through shape functions $s_\lambda(t), s_\beta(t) \in [0,1]$:
\begin{equation}\label{eq:lambda_beta_schedule}
  \lambda(t) \;=\; \lambda_\text{max} \cdot s_\lambda(t), \qquad
  \beta(t) \;=\; 2 - (2 - \beta_\text{min}) \cdot s_\beta(t).
\end{equation}
The shape functions control how strongly the objective departs from the regression-like limit: $s_\lambda(t)=0$ removes the interaction term, $s_\beta(t)=0$ gives $\beta(t)=2$, and $\beta_\text{min}=1$ gives the standard energy score at the schedule peak. We use five profile families (Appendix~\ref{app:on_lambda_beta}): \emph{constant} (the original DDM), \emph{linear}, \emph{step}, \emph{SNR}, and \emph{dynamical-regimes} (\emph{dyn-reg}), which is piecewise linear in $t$ with $s(t)=1$ for $t\le t_s$ and $s(t)=0$ for $t\ge t_\text{sep}$. By default, we use the linear profile $s_\lambda(t)=s_\beta(t)=1-t$ with $\beta_{\min}=0.1$, which matches dyn-reg (Tab.~\ref{tab:schedule_profile}) without requiring the regime thresholds.

\vspace{-0.1em}
\section{Experiments}

\vspace{-0.1em}
\label{sec:experiments}

The section is organized as follows. We first ablate our design choices (Tab.~\ref{tab:main_ablation}), then compare against few-step methods in a controlled setting (Tab.~\ref{tab:controlled_comparison}) and finally at system level at large scale (Tab.~\ref{tab:main_comparison}).

\vspace{-0.1in}
\paragraph{Experiment Settings.}
We train B-scale latent DiTs (depth 12, width 768)~\citep{peebles2023dit,vaswani2017attention,dosovitskiy2021an,rombachHighResCVPR} on class-conditional ImageNet-$256^2$~\citep{imagenet} for 400k steps using AdamW~\citep{loshchilov2018decoupled} unless noted otherwise (all hyperparameters in~\cref{tab:training_setup}). Evaluations primarily use FID@50k$\downarrow$~\citep{heusel2017gans} in many- (50) and few-step (4) settings, with individually swept CFG scales~\citep{ho2022cfg}. We report FD-DINOv2, KID, KDD, Inception Score, precision, and recall in Appendix ~\ref{app:additional_results}.

\begin{table}[t]
    \centering
    \caption{\textbf{Main Ablation.} Starting from DDM~\citep{debortoli2025distributional} baseline (DiT-B, ImageNet-$256^2$), we gradually add our proposed components.  (\textsc{Config} \textbf{F}) achieves the best performance.
    }
    \vspace{.2em}
    \scalebox{.85}{
    \begin{tabular}{c@{\hskip .5em}lccccc}
        \toprule
        \multicolumn{2}{l}{\multirow{2}{*}[-.2em]{Method}} & \multirow{2}{*}[-.2em]{Train Cost} & \multicolumn{4}{c}{FID$\downarrow$ (by steps)} \\
        \cmidrule(lr){4-7}
        & & & 50 & 20 & 10 & 4 \\
        \midrule
        {\color{ourgray}--} & {\color{ourgray} FM Reference} & 1.0$\times$ & {\color{ourgray}4.97} & {\color{ourgray}5.64} & {\color{ourgray}8.11} & {\color{ourgray}26.53} \\
        \textbf{A} & DDM Baseline (na\"ive adaptation to transformers) & 4.0$\times$ & 10.04 & 10.16 & 11.85 & 43.68 \\
        \textbf{B} & + Deferred Expansion & 1.5$\times$ & 5.41 & 5.71 & 7.25 & 17.51 \\
        \textbf{C} & + $t$-Adaptive Noise Gating & 1.5$\times$ & 5.50 & 5.70 & 7.02 & 16.02 \\
        \textbf{D} & + Improved $\xi$ Conditioning & 1.5$\times$ & 5.47 & 5.62 & 6.97 & 15.74 \\
        \textbf{E} & + Time-dependent Scoring Rule Schedules & 1.5$\times$ & 5.69 & 5.98 & 7.11 & 14.33 \\
        \textbf{F} & + Improved Train-time $t$ Sampling & 1.5$\times$ & \textbf{4.56} & \textbf{4.98} & \textbf{6.42} & \textbf{13.13} \\ 
        \bottomrule
    \end{tabular}
    }
    \label{tab:main_ablation}
\end{table}

\vspace{-0.05em}
\subsection{Ablation Study}
\label{sec:ablation}
\vspace{-0.05em}

We ablate the design decisions of Section~\ref{sec:method} along three axes: (i) architectural choices, including deferred particle expansion (Sec~\ref{sec:efficient_training}), $t$-dependent $\xi$ gating (App~\ref{app:xi_ablation}), and noise conditioning variants (Fig~\ref{fig:xi_conditioning}); (ii) time-dependent scoring rule schedules $(\lambda(t),\beta(t))$ (Sec~\ref{sec:time_dependent_scoring_rule_schedules}); and (iii) the training time distribution $p(t)$ in~\eqref{eq:ddm_loss}. \Cref{tab:main_ablation} adds them step by step, from a DDM~\citep{debortoli2025distributional} baseline (\textsc{Config}~\textbf{A}) to our final method (\textsc{Config}~\textbf{F}). To disentangle individual contributions, \cref{tab:independent_component_ablation} adds each independently to \textsc{Config}~\textbf{B}, and all improve the 4-step result without retuning.

\subsubsection{Deferred Expansion \& Architectural Improvements}\label{sec:ablation_architectural_improvements}

\begin{figure}[t]
    \centering
    \includegraphics[scale=.6]{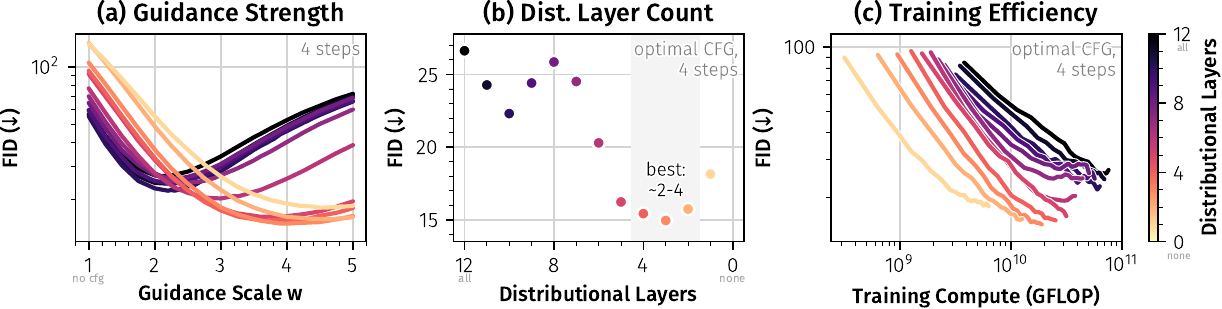}
    \caption{\textbf{Few-Step Generation Quality vs.~Number of Distributional Layers.} We train a DiT-B on ImageNet for 400k steps and evaluate FID on 50k 4-step samples, optionally using CFG. \textbf{(a)} In the unguided setting, more distributional layers improve generation quality. However, this trend \emph{reverses in the practical setting} with CFG, with models with \emph{fewer} distributional layers performing better. \textbf{(b)} Under optimal guidance, 2-4 distributional layers lead to to the best results. \textbf{(c)} Training efficiency decreases significantly with many distributional layers, but convergence bottoms out early with too few (notably, also for some models with higher distributional layer count). Similarly to the step-matched setting in (b), around 2-5 distributional layers offer a good efficiency-quality trade-off.
    }
    \label{fig:deferred_ablation_plots}
    \vspace{-0.2in}
\end{figure}

In \cref{tab:main_ablation}, we gradually add deferred expansion and architectural improvements (\textsc{Configs} \textbf{A}$\rightarrow$\textbf{D}). Then in \cref{tab:architecture_params_ablation} and \cref{fig:deferred_ablation_plots}, we report results starting from \textsc{Config} \textbf{D} to account for joint interactions.

\vspace{-0.1in}
\paragraph{Deferred Expansion (\textsc{Config} \textbf{B}).}
Compared to a FM~\citep{lipman2023flow} baseline, DDMs~\citep{debortoli2025distributional} incur a large training cost overhead proportional to the number of particles $m$ \footnote{The cost of the forward-backward pass scales \emph{linearly} with $m$ while the loss computation scales \emph{quadratically} with $m$. We only focus on the forward-backward cost since it is the costliest operation in the training loop.}.

\textit{How much should expansion be deferred?}
In \cref{fig:deferred_ablation_plots,tab:deferred_expansion_hparam_ablation}, we vary the number of distributional layers.
Without guidance (\cref{fig:deferred_ablation_plots}a, $w = 1$), less deferral leads to better performance, leading to a direct trade-off between training cost and generation quality.
However, in practical settings, CFG~\citep{ho2022cfg} is typically applied to obtain better generation quality. Once applied, the behavior is reversed: models with \emph{fewer} distributional layers tolerate high guidance scales, enabling \emph{improved performance} under guidance (\cref{fig:deferred_ablation_plots}b).
This leads to \emph{simultaneous} reductions in training costs (to $1.5\times$ of FM cost compared to $4\times$ of baseline DDM) and improvements in generation quality.

\vspace{-0.1in}
\paragraph{Timestep-Adaptive Noise Gating (\textsc{Config} \textbf{C}).}

We study the impact of a learned $t$-dependent gating mechanism~\eqref{eq:adaptive_ksi_gating} which controls the magnitude of noise $\xi$ (see App.~\ref{app:xi_ablation}). This slightly reduces many-step performance but boosts few-step generation quality (Tab~\ref{tab:main_ablation} \textbf{B}$\rightarrow$\textbf{C}; Tab~\ref{tab:noise_gating_hparam_ablation}).

\vspace{-0.1in}
\paragraph{Improved $\boldsymbol{\xi}$ Conditioning (\textsc{Config} \textbf{D}).}
Compared to DDM~\citep{debortoli2025distributional}, which concatenates $\xi$ channel-wise to the model input in pixel space, our deferred expansion affords a larger design space, explored in Tab.~\ref{tab:xi_conditioning_hparam_ablation}. Among concatenation variants, full widening expands all late-layer projections, fixed-inner-width widening carries $\xi$ only in the residual stream, and substitution overwrites channels. Fixed-inner-width concatenation gives the best trade-off, so we adopt it. Adaptive norms and token conditioning underperform \emph{severely}, especially with few steps.

\begin{table}[t]
    \centering
    \caption{\textbf{Architecture Design Space.} Starting from \textsc{Config} \textbf{D} (\cref{tab:main_ablation}), we explore optimal hyper\-parameters for our architectural changes using FID$\downarrow$. \textbf{Best}, \underline{2nd}, and \ \boxhighlight{gray!12}{final choices}\ \ are highlighted.
    \vspace{-0.1in}
    }
    \label{tab:architecture_params_ablation}
    \newcommand{\tabscale}[1]{\scalebox{.75}{#1}}
    \newcommand{\normal}[1]{\multicolumn{1}{c}{#1}}
    \let\oldT\T
    \renewcommand{\T}[0]{\normal{\oldT{}}}
    \newcommand{\best}[1]{\multicolumn{1}{c}{\textbf{\roundtwofixed{#1}}}}
    \newcommand{\second}[1]{\multicolumn{1}{c}{\underline{\roundtwofixed{#1}}}}
    \newcommand{\bestt}[1]{\bfseries\boldmath\textbf{#1}}
    \newcommand{\secondt}[1]{\underline{#1}}
    \setlength{\tabcolsep}{.3em} 
    \newcommand{\chosenparam}{\rowcolor{gray!12}}
    \newcommand{\chosencell}{\cellcolor{gray!12}}
    \newcommand{\chosennum}[1]{\multicolumn{1}{>{\columncolor{gray!12}}c}{\roundtwofixed{#1}}}
    \newcommand{\secondchosen}[1]{\multicolumn{1}{>{\columncolor{gray!12}}c}{\underline{\roundtwofixed{#1}}}}
    \newcommand{\unchosen}[1]{\cellcolor{white}#1}
    \newcommand{\unchosenR}[3]{%
        \multicolumn{1}{>{\columncolor{white}}R{#1}{#2}}{#3}%
    }
    \hfill
    \autosubtable{\textbf{Expansion Start Layer.} Few-step inference performs best for $\ell_\text{start}\in [8,10]$. We choose $\ell_\text{start} = 10$ as it combines good few-step results with low extra cost.}{tab:deferred_expansion_hparam_ablation}{\tabscale{
        \begin{tabular}{lcR{1}{2}R{2}{2}|lcR{1}{2}R{2}{2}}
            \toprule
            $\ell_\text{start}$ & Cost & \normal{50-step} & \normal{4-step} & $\ell_\text{start}$ & Cost & \normal{50-step} & \normal{4-step} \\
            \midrule
            0  & 4.00$\times$ & 7.670 & 26.624 & 6  & 2.50$\times$ & 6.386 & 20.267 \\
            1  & 3.75$\times$ & 7.791 & 24.272 & 7  & 2.25$\times$ & 6.296 & 16.233 \\
            2  & 3.50$\times$ & 7.390 & 22.207 & 8  & 2.00$\times$ & 5.554 & \second{15.429} \\
            3  & 3.25$\times$ & 7.028 & 24.406 & 9  & 1.75$\times$ & 5.592 & \best{14.956} \\
            4  & 3.00$\times$ & 6.99 & 25.85 & \chosencell 10 & \chosencell\secondt{1.50$\times$} & \secondchosen{5.470} & \chosennum{15.741} \\
            5  & 2.75$\times$ & 6.613 & 24.511 & 11 & \bestt{1.25$\times$} & \best{5.407} & 18.150 \\
            \bottomrule
        \end{tabular}
    }}
    \hfill
    \hspace{0.1em}
    \hfill
    \autosubtable{\textbf{$\boldsymbol{\xi}$ Conditioning.} Ada-norms/$\xi$ tokens work poorly; fixed concat works best.}{tab:xi_conditioning_hparam_ablation}{\tabscale{
        \begin{tabular}{lR{1}{2}R{2}{2}}
            \toprule
            Mechanism & \normal{50-step} & \normal{4-step} \\
            \midrule
            ada-norm   & 6.625 & 97.063 \\
            additive   & 5.488 & 16.996 \\
            tokens     & 6.379 & 70.169 \\
            concat subst. & 5.505 & \second{16.024} \\
            \chosenparam concat+fixed  & \second{5.470} & \best{15.741} \\
            concat+scale\rlap{\smash{$^\dagger$}}  & \best{5.230} & 16.287 \\
            \bottomrule
        \end{tabular}
    }}
    \hfill
    \hspace{0.1em}
    \hfill
    \autosubtable{\textbf{Noise Gating.} Learn\-ed $t$-adaptive gating improves few-step results.}{tab:noise_gating_hparam_ablation}{\tabscale{
        \begin{tabular}{lR{1}{2}R{2}{2}}
            \toprule
            Gating & \normal{50-step} & \normal{4-step} \\
            \midrule
            static & \best{5.346} & \second{17.465} \\
            \chosenparam $t$-adaptive & \second{5.470} & \best{15.741} \\
            \bottomrule
            \\[.1em]
            \multicolumn{3}{c}{\makebox[0pt][c]{\shortstack{$^\dagger$adds substantial additional\\\ trainable parameters ($\sim$10M)}}}
        \end{tabular}
    }}
    \hfill
    \vspace{-0.1in}
\end{table}

\subsubsection{Training Objective Improvements}\label{sec:ablation_train_objective}

We explore training improvements: time-dependent schedules $(\lambda(t),\beta(t))$ (Section~\ref{sec:time_dependent_scoring_rule_schedules}), the training time distribution $p(t)$, and kernel spatial modes (see Appendix~\ref{app:kernel_modes}). The results are in \cref{tab:train_objective_params_ablation}.

\begin{table}[!t]
    \centering
    \caption{
    \textbf{Training Objective Design Space.} We explore the optimal choice of hyperparameters and ablate our training objective improvements, using FID$\downarrow$. \textbf{Best}, \underline{2nd}, and \ \boxhighlight{gray!12}{\vphantom{;}final choices}\ \ are highlighted.
    \vspace{-0.2in}
    }
    \label{tab:train_objective_params_ablation}
    \newcommand{\chosenparam}[0]{\rowcolor{gray!12}}
    \newcommand{\tabscale}[1]{\scalebox{.75}{#1}}
    \newcommand{\lefttabscale}[1]{\scalebox{.65}{#1}}
    \newcommand{\normal}[1]{\multicolumn{1}{c}{#1}}
    \let\oldT\T
    \renewcommand{\T}[0]{\normal{\oldT{}}}
    \newcommand{\best}[1]{\multicolumn{1}{c}{\textbf{\roundtwofixed{#1}}}}
    \newcommand{\second}[1]{\multicolumn{1}{c}{\underline{\roundtwofixed{#1}}}}
    \newcommand{\bestp}[1]{\multicolumn{1}{c}{\phantom{0}\textbf{\roundtwofixed{#1}}}}
    \newcommand{\secondp}[1]{\multicolumn{1}{c}{\phantom{0}\underline{\roundtwofixed{#1}}}}
    \newcommand{\bestt}[1]{\bfseries\boldmath\textbf{#1}}
    \newcommand{\secondt}[1]{\underline{#1}}
    \setlength{\tabcolsep}{.3em} 
    \newcommand{\topaligned}[1]{\raisebox{-\height}[0pt][\totalheight]{#1}}
    \newcommand{\leftfirstcol}[1]{\makebox[4.5em][l]{#1}}
    \hfill
    \topaligned{\begin{tabular}[t]{@{}c@{}}
        \autosubtable{\textbf{Initial Exponent.}\vspace{-.65em}}{tab:initial_exponent}{\lefttabscale{
            \begin{tabular}{lR{1}{2}R{2}{2}}
                \toprule
                \leftfirstcol{$\beta(t=0)$} & \normal{50-step} & \normal{4-step} \\
                \midrule
                1.00 & \best{5.340} & 14.801 \\
                0.50 & \second{5.432} & 15.041 \\
                \chosenparam 0.10 & \normal{5.69} & \best{14.334}\\
                \rowcolor{white} 0.01 & 5.885 & \second{14.415} \\
                \bottomrule
            \end{tabular}
        }}\\\noalign{\vskip .18em}
        \autosubtable{\setcounter{subtable}{3}\textbf{Train $t$ Sampling.}\vspace{-.65em}}{tab:train_t_sampling}{\lefttabscale{
            \begin{tabular}{lR{1}{2}R{2}{2}}
                \toprule
                \leftfirstcol{$p(t)$} & \normal{50-step} & \normal{4-step} \\
                \midrule
                imf     & \second{5.178} & \second{13.980} \\
                \chosenparam jit     & \best{4.568} & \best{13.133} \\
                \rowcolor{white} logit   & 5.704 & 14.584 \\
                uniform & 6.608 & 15.391 \\
                \bottomrule
            \end{tabular}
        }}\\\noalign{\vskip .18em}
        \autosubtable{\setcounter{subtable}{4}\textbf{Kernel sp. mode.}\vspace{-.65em}}{tab:kernel}{\lefttabscale{
            \begin{tabular}{lR{1}{2}R{2}{2}}
                \toprule
                \leftfirstcol{Kernel} & \normal{50-step} & \normal{4-step} \\
                \midrule
                global & \second{5.513} & \second{18.356} \\
                \chosenparam local  & \best{5.418} & \best{15.884} \\
                \bottomrule
            \end{tabular}
        }}
    \end{tabular}}
    \hfill
    \topaligned{\setcounter{subtable}{1}\autosubtable{\textbf{Scoring Rule Scheduling.} (i) Linear scheduling gives the best few-step results. Note however that the landscape changes as the number of steps increase. For 50-steps, step combination, with early switch to diffusion-like behavior dominate (ii-iii) We also explore omitting either scheduler. While most of those combinations underperform, we find surprisingly that keeping $\lambda=1$ and a step schedule for $\beta$ yields good results.}{tab:schedule_profile}{\tabscale{
        \begin{tabular}{lR{2}{2}R{2}{2}}
            \multicolumn{3}{c}{(i) \textbf{Schedule $\beta(t),\lambda(t)$.}} \\[.3em]
            \toprule
            Sched.~Profile & \normal{50-step} & \normal{4-step} \\
            \midrule
            \chosenparam linear & \normal{\phantom{0}5.69} & \second{14.334} \\
            dyn-reg           & 5.727 & \best{14.272} \\
            snr                 & 6.423 & 14.640 \\
            step-0.90           & 7.935 & 19.937 \\
            step-0.75           & 12.284 & 25.163 \\
            step-0.50           & 8.765 & 20.175 \\
            step-0.25           & \secondp{5.418} & 16.300 \\
            step-0.10           & \bestp{4.907} & 16.233 \\
            \bottomrule
        \end{tabular}
        \begin{tabular}{lR{2}{2}R{2}{2}}
            \multicolumn{3}{@{\hskip -1em}c@{\hskip -1em}}{(ii) \textbf{Ablation:} only $\beta(t)$, $\lambda=1$} \\[.3em]
            \toprule
            Sched.~Profile & \normal{50-step} & \normal{4-step} \\
            \midrule
            linear    & 6.120 & 16.902 \\
            dyn-reg & 6.343 & 16.663 \\
            snr       & 6.955 & 17.710 \\
            step-0.90 & 7.945 & 22.313 \\
            step-0.75 & 12.635 & 28.582 \\
            step-0.50 & 9.208 & 18.350 \\
            step-0.25 & \secondp{5.607} & \second{14.377} \\
            step-0.10 & \bestp{5.096} & \best{13.063} \\
            \bottomrule
        \end{tabular}
        \begin{tabular}{lR{1}{2}R{2}{2}}
            \multicolumn{3}{@{\hskip -1em}c@{\hskip -1em}}{(iii) \textbf{Ablation:} only $\lambda(t)$, $\beta=1$} \\[.3em]
            \toprule
            Sched.~Profile & \normal{50-step} & \normal{4-step} \\
            \midrule
            linear    & 5.348 & 14.869 \\
            dyn-reg & 5.347 & \best{13.600} \\
            snr       & \best{5.242} & 14.257 \\
            step-0.90 & 5.457 & 16.487 \\
            step-0.75 & 5.620 & \second{14.072} \\
            step-0.50 & 5.379 & 14.110 \\
            step-0.25 & \second{5.280} & 15.075 \\
            step-0.10 & 5.736 & 17.309 \\
            \bottomrule
        \end{tabular}
    }}}
    \hfill
    \hfill
    \hfill
    \vspace{-0.15in}
\end{table}

\vspace{-0.1in}
\paragraph{Time-dependent Scoring Rule Schedulers (\textsc{Config} \textbf{E}).}
\label{sec:time_dependent_scoring_rules_exp} Following section~\ref{sec:time_dependent_scoring_rule_schedules}, we use time-dependent schedules for $(\lambda(t),\beta(t))$ -- hyperparameters of the generalized energy score~\eqref{eq:kernel_score}. The results for many options of different schedules are given in Table~\ref{tab:train_objective_params_ablation}b.
The linear schedule overall leads to the best trade off in performance for 50 and 4 steps regime. The rest of the results in Table~\ref{tab:train_objective_params_ablation} are using this linear schedule.

\textit{Do both the norm exponent $\beta$ and the interaction weight $\lambda$ require scheduling?}
We also explore omitting the scheduler for either $\beta$ or $\lambda$ in \cref{tab:schedule_profile}-ii and \cref{tab:schedule_profile}-iii. While most combinations underperform, scheduling only $\beta$ with $\lambda=1$ and an early switch at $t=0.1$ yields particularly strong FID. We do not adopt this setting as our default because at $(\lambda,\beta)=(1,2)$ the score reduces to $\|\mu - y\|^2$, independent of the population covariance, so the spread of the model's samples is left unconstrained for $t>0.1$. Under our schedules, where $\lambda<1$, the variance remains identified. A theoretically optimal choice of $\lambda(t)$ and $\beta(t)$ remains an open question.

\textit{How to set the initial norm exponent $\beta(t=0)$?} The schedule for $\beta(t)$ is an increasing function of $t$ with the lowest value at $\beta(t=0)$. In \cref{tab:initial_exponent} we run the ablation over this value and find that a value of $0.1$ yields the best trade-off between different sampling budgets.

\vspace{-0.1in}
\paragraph{Improved Train-time $t$ Sampling (\textsc{Config} \textbf{F}).}
Finally, \cref{tab:train_t_sampling} compares different distributions from which to sample $t$ during training.
Since we apply no further $t$-dependent loss weighting, these control both the implicit time-dependent weighting of the objective and the distribution of interpolants $x_t$ seen during training. We observe that \texttt{jit} distribution leads to the best results. This noise distribution biases the time towards the initial noise, where distributional losses matter more. 

For compactness, we report FID only in the main text. The improvements persist across all additional metrics, including FD-DINOv2, KID, KDD, Inception Score, precision, and recall and are reported in~\cref{tab:all_metrics}. Repeated sampling from fixed $x_t$ further confirms that iDDM produces conditional diversity that deterministic FM cannot (\cref{tab:cond-div}). Accounting for iDDM's higher per-step training cost, a compute-matched comparison shows the same few-step advantage: at the training budget of the 400k-step FM baseline, iDDM (260k steps) improves 4-step FID from 26.70 to 14.33, while FM remains slightly better at 50 steps (4.97 vs.\ 5.62); see~\cref{tab:compute_scaling} for the full compute sweep.

\vspace{-0.05em}
\subsection{Comparison with Previous Methods}
\vspace{-0.05em}

\paragraph{Controlled Comparison on ImageNet-256$^2$.}
Existing few-step methods~\citep{zhou2025inductive,geng2025mean,geng2026improved,potaptchik2026metaflowmapsenable} typically report results in highly varying settings, making fair comparisons challenging.
We therefore compare with them in a controlled setting, retraining all models from scratch at B-scale with comparable settings. We re-sweep relevant hyperparameters to ensure a fair comparison.
As shown in \Cref{tab:controlled_comparison}, iDDM substantially improves over naïve DDM, reducing 4-step FID from 43.68 to 13.13 and 50-step FID from 10.04 to 4.57, and the deterministic baselines, while remaining competitive with stochastic flow-map methods across sampling budgets.

\vspace{-0.1in}

\paragraph{System-Level Results.} At XL scale, iDDM-XL/2 reaches 2.38 FID at 50 steps and 4.48 at 4 steps after 200 epochs (samples in~\cref{fig:teaser}). \Cref{tab:main_comparison} compares iDDM against deterministic few-step flow-map methods trained from scratch~\citep{frans2025one,geng2025mean,geng2026improved,zhou2025inductive} and against MFM~\citep{potaptchik2026metaflowmapsenable}, which uses multi-stage training from a pretrained model. Entries marked $^*$ are taken from the respective papers, all others are evaluated by us with official checkpoints and code. Both DDM and iDDM are trained from scratch.

\vspace{-0.1in}
\paragraph{Positioning w.r.t.\ MeanFlow-style and few-step methods.}
At 1-2 steps, teacher-based distillation such as sCD~\citep{lu2025simplifying} and ADD~\citep{sauer_distillation} achieves the lowest FID, but targets a different operating point: it requires a pretrained many-step teacher and a second training stage while iDDM doesn't. Among methods trained from scratch, published iMF and IMM checkpoints attain lower FID than iDDM at every inference budget, at a substantially higher training cost: iMF uses a 48-layer model and roughly $17.6\times$ our total training budget, and IMM $19\times$ more training epochs. iDDM instead combines three properties that no baseline in \cref{tab:imagenet_256_comparison} offers together. (i)~It is \emph{stochastic}: it learns a conditional predictor trained to sample from $p(x_1 \mid x_t)$, whereas MeanFlow, iMF, and IMM learn deterministic maps (see App.~\ref{app:posterior-div}, par. "Posterior Diversity" and Tab.~\ref{tab:cond-div} for a comparison with FM). (ii)~It is trained from scratch in a \emph{single stage}, whereas the published MFM results rely on a pretrained flow map and multi-stage training. (iii)~A \emph{single checkpoint} serves the full 4-50 step range without degrading: its FID is non-increasing in the number of steps (4.48 $\to$ 2.38), whereas MeanFlow (2.93 $\to$ 3.29) and MFM (1.97 $\to$ 5.74) are stronger at 4-8 steps but get worse with additional sampling compute. iDDM is also relatively inexpensive to train: per training step it costs $1.41\times$ flow matching, versus $1.96\times$ for MeanFlow and $6.20\times$ for iMF (\Cref{tab:training_efficiency}). Finally, against flow matching at matched training compute (\cref{tab:compute_scaling}), iDDM halves 4-step FID (13.13 vs.\ 26.36) while staying within 0.40 FID at 50 steps (4.61 vs.\ 4.21).

\begin{figure}[htbp]
\centering
\begin{tabular}{@{}c@{}c@{}c@{}c@{}c@{}}
\includegraphics[width=0.2\textwidth]{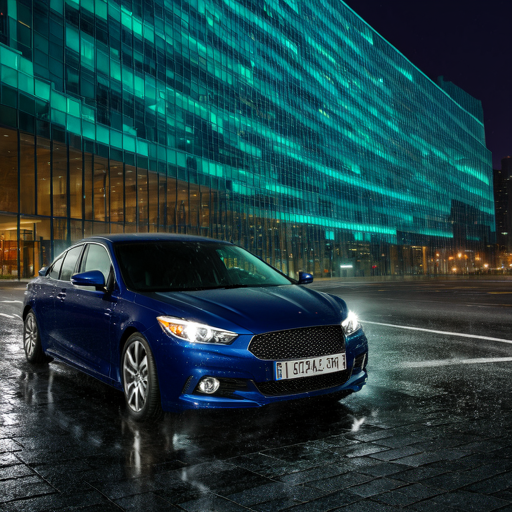} &
\includegraphics[width=0.2\textwidth]{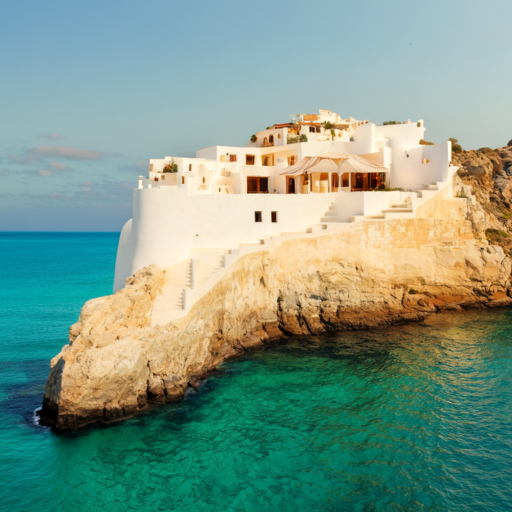} &
\includegraphics[width=0.2\textwidth]{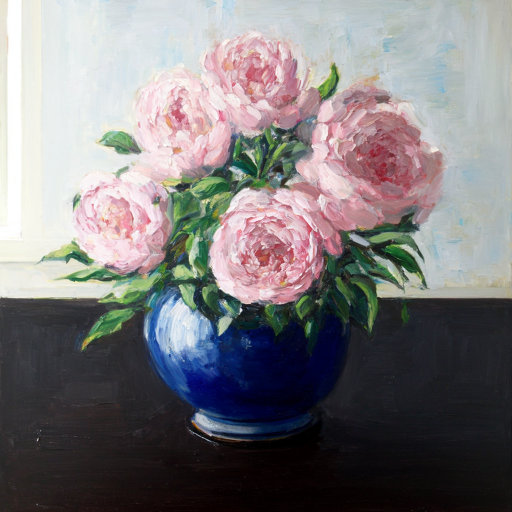} &
\includegraphics[width=0.2\textwidth]{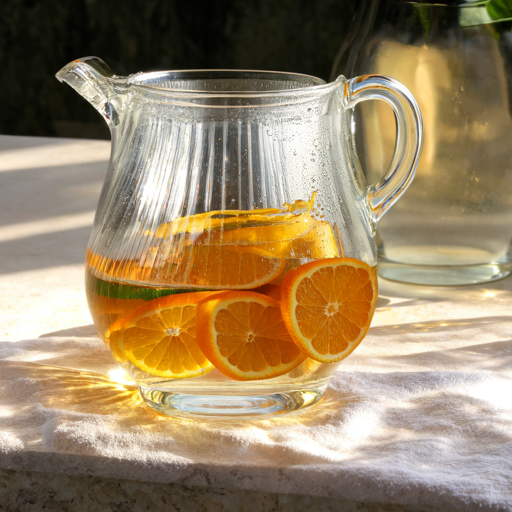} &
\includegraphics[width=0.2\textwidth]{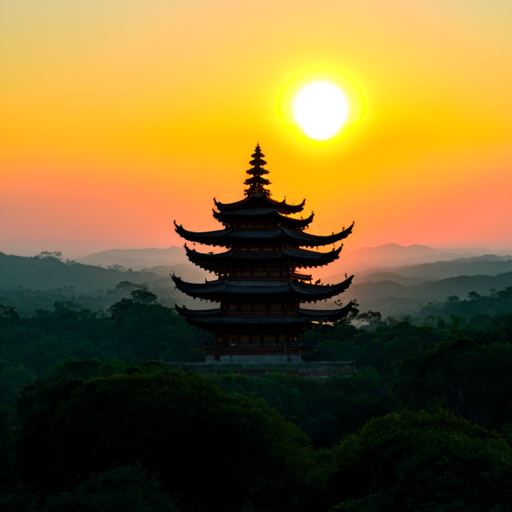} \\
\includegraphics[width=0.2\textwidth]{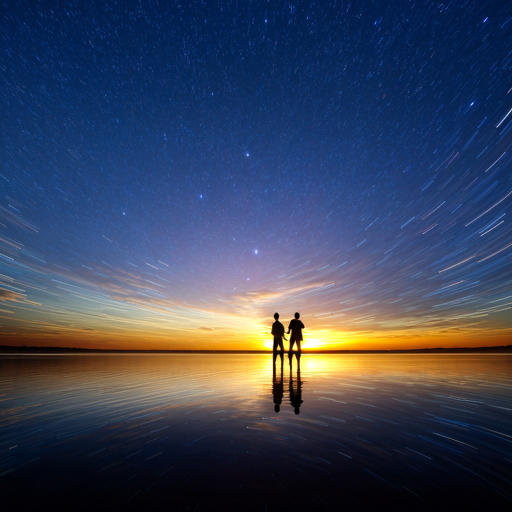} &
\includegraphics[width=0.2\textwidth]{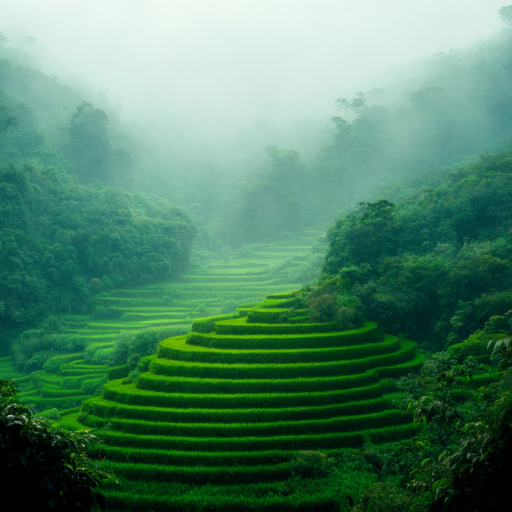} &
\includegraphics[width=0.2\textwidth]{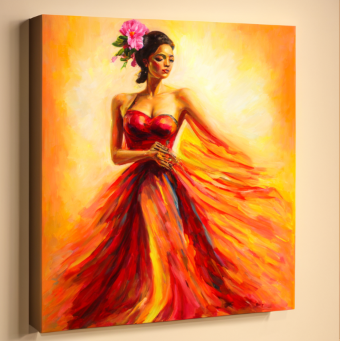} &
\includegraphics[width=0.2\textwidth]{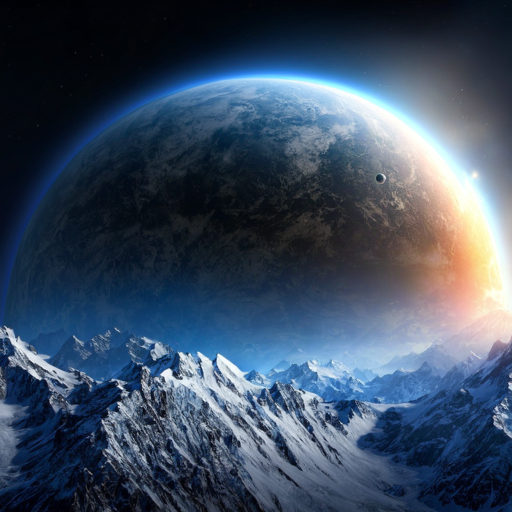} &
\includegraphics[width=0.2\textwidth]{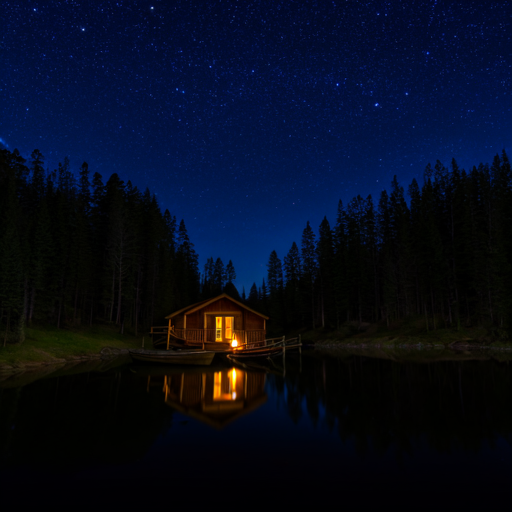}
\end{tabular}
\caption{Example images generated by our text-to-image model at 512$^2$ resolution. For this, we finetune for 20k training steps our model pretrained on 256$^2$ resolution.}
\label{fig:text2image-grid}
\end{figure}

\vspace{-0.1in}
\paragraph{Transfer to T2I.} We transfer iDDM to a 1.6B-parameter text-to-image model. Relative to a matched FM baseline, iDDM degrades far less as the budget decreases, improving MS-COCO FID from 78.20 to 41.05 at 4 steps and from 26.33 to 19.40 at 8 steps (App.~\ref{app:text_to_image}). Samples in~\cref{fig:text2image-grid}.

\begin{table}[t]
    \centering
    \caption{
    \textbf{Few-Step Comparison on ImageNet-256$^\mathbf{2}$} (FID$\downarrow$). iDDM \emph{significantly} improves upon DDM and achieves competitive FID across all steps while training from scratch in a single stage.
    \vspace{-0.1in}
    }
    \label{tab:imagenet_256_comparison}
    \newcommand{\chosenparam}[0]{\rowcolor{gray!12}}
    \newcommand{\tabscale}[1]{\scalebox{.71}{#1}}
    \newsavebox{\visscalebox}
    \newcommand{\viswiden}[2][1.05]{%
      \sbox{\visscalebox}{#2}%
      \makebox[\wd\visscalebox][c]{%
        \raisebox{\dimexpr\ht\visscalebox-#1\ht\visscalebox\relax}[%
          \ht\visscalebox][\dp\visscalebox]{%
          \scalebox{#1}{\usebox{\visscalebox}}%
        }%
      }%
    }
    \newcommand{\normal}[1]{\multicolumn{1}{c}{#1}}
    \newcommand{\best}[1]{\multicolumn{1}{c}{\textbf{\roundtwofixed{#1}}}}
    \newcommand{\second}[1]{\multicolumn{1}{c}{\underline{\roundtwofixed{#1}}}}
    \newcommand{\bestp}[1]{\multicolumn{1}{c}{\phantom{0}\textbf{\roundtwofixed{#1}}}}
    \newcommand{\secondp}[1]{\multicolumn{1}{c}{\phantom{0}\underline{\roundtwofixed{#1}}}}
    \newcommand{\bestt}[1]{\bfseries\boldmath\textbf{#1}}
    \newcommand{\secondt}[1]{\underline{#1}}
    \newcommand{\stepheader}[1]{{\hskip -.2em}#1{\hskip -.2em}}
    \setlength{\tabcolsep}{.3em} 
    \newcommand{\topaligned}[1]{\raisebox{-\height}[0pt][\totalheight]{#1}}
    \newcommand{\bottomaligned}[1]{\raisebox{\depth}{#1}}
    \bottomaligned{\autosubtable{\textbf{Controlled Comparison} at DiT-B scale (depth 12, width 768), retrained from scratch under same standard settings (400k steps/80ep).}{tab:controlled_comparison}{\scalebox{.8}{%
        \begin{tabular*}{.4\textwidth}{@{\extracolsep{\fill}}lcc}
            \toprule
            Method & \stepheader{50-step} & \stepheader{4-step} \\
            \midrule
            \multicolumn{3}{l}{\textit{\textbf{Deterministic Methods}}} \\
            Shortcut & \phantom{0}9.85\rlap{$^\S$} & 15.02 \\
            IMM & \phantom{0}7.57 & 24.85 \\
            \multicolumn{3}{l}{\textit{\textbf{Stochastic Methods}}} \\
            MFM & \phantom{0}7.72 & \bestp{9.17} \\
            DDM (na\"ive) & 10.04 & 43.68 \\
            iDDM (ours) & \bestp{4.568} & 13.13 \\
            \bottomrule
        \end{tabular*}%
    }}}
    \hfill
    \bottomaligned{\autosubtable{\textbf{System-level Comparison} at DiT-XL scale (depth 28, width 1152). $^\dagger$iMF uses a much deeper transformer (28$\rightarrow$48 layers).}{tab:main_comparison}{\scalebox{.8}{%
        \begin{tabular*}{.8125\textwidth}{@{\extracolsep{\fill}}lccccc}
            \toprule
            Method & \#Epochs & \stepheader{50-step} & \stepheader{16-step} & \stepheader{8-step} & \stepheader{4-step} \\
            \midrule
            \multicolumn{6}{l}{\textit{\textbf{Deterministic Few-step Flow-map models from scratch}}} \\
            Shortcut-XL/2 & \phantom{0}162 & 4.68\rlap{$^\S$} & 5.39 & 5.74 & 7.8\phantom{0}\rlap{$^{\ast}$} \\
            MeanFlow-XL/2 & \phantom{0}240 & 3.29 & 3.06 & 2.93 & 2.93 \\
            iMF-XL/2$^{\dagger}$ & \phantom{0}800 & 1.43 & 1.43 & 1.50 & 1.51 \\
            IMM-XL/2 & 3840 & 1.82 & 1.86 & 1.99\rlap{$^{\ast}$} & 2.51\rlap{$^{\ast}$} \\
            \multicolumn{6}{l}{\textit{\textbf{Stochastic Few-step Flow-map (multi-stage training)}}} \\
            MFM-XL/2 & 800+80+28 & 5.74 & 3.29 & 2.19 & 1.97\rlap{$^{\ast}$} \\
            \multicolumn{6}{l}{\textit{\textbf{Stochastic Few-step diffusion/FM from scratch}}} \\
            DDM-XL/2 (na\"ive) & \phantom{0}200 & 4.71 & 4.50 & 5.70 & 23.25 \\
            iDDM-XL/2 (ours) & \phantom{0}200 & 2.38 & 3.03 & 3.04 & 4.48 \\
            \bottomrule
        \end{tabular*}%
    }}}
    \par\vspace{.4em}
    {\footnotesize $^\S$evaluated at 64 sampling steps since Shortcut only supports powers of two.}
\end{table}

\vspace{-0.1em}
\section{Related Work}
\label{sec:related_work}

\vspace{-0.05in}
\paragraph{Accelerating diffusion sampling.}
Distillation~\citep{luhman2021knowledge, sauer_distillation, Yin_2024_CVPR, salimans2022progressive, liu2023instaflow, pmlr-v202-song23a, xu2024ufogen, salimans2024multistep} trains a few-step student against a many-step teacher; improved samplers~\citep{lu2022dpmsolver, Lu_2025, zheng2023dpmsolverv, karras2022elucidating} reduce discretization error of a frozen model; parallel and speculative methods~\citep{shih2023parallel, chen2024accelerating, bortoli2025accelerated} address the sequential nature of diffusion models. Our work is complementary to all three: we modify the training objective, not the sampler. For instance, our method integrates with the Universal Inverse Distillation framework of \citet{kornilov2025universal} by swapping the diffusion loss for a distributional one. 

\vspace{-0.1in}
\paragraph{Learning the diffusion posterior.}
Several approaches enrich the standard mean-prediction denoiser with (co)variance information for few-step sampling~\citep{NEURIPS2020_4c5bcfec, pmlr-v139-nichol21a, pmlr-v162-bao22d, bao2022analyticdpm, ou2025improving}. Diffusion-GAN~\citep{xiao2022tackling} and UFOGen~\citep{xu2024ufogen} learn multimodal conditional generators via adversarial training conditioned on a noise variable $\xi$---architecturally close to DDMs but trained against a discriminator instead of a proper scoring rule. \citet{aiello2024fast} and \citet{galashov2025deep} use MMD on CLIP features and MMD gradient flows respectively; \citet{wang2025diffusedisperseimagegeneration} regularises representations for diversity without explicit distributional training. We build directly on the generalized energy score formulation of \citep{debortoli2025distributional,shen2025reverse}, addressing its training cost and its static hyperparameters. 
Very recent works also investigate the use of \emph{distributional losses} in representation space \citet{yang2026representation}. 
Extended related work on the distributional lineage of our method can be found in Section \ref{sec:mmd-lineage}.

\vspace{-0.1in}
\paragraph{Consistency, moment-matching, and fastforward methods.}
Consistency Models~\citep{pmlr-v202-song23a, song2024improved} and extensions like Consistency Trajectory Models~\citep{kim2024consistency} and Shortcut Models~\citep{frans2025one} learn self-consistent few-step generators. MeanFlow~\citep{geng2025mean} learns an average velocity field for one-step generation, and the improved iMF variant addresses training stability and recovers inference flexibility ~\citep{geng2026improved}. AlphaFlow~\citep{zhang2026alphaflow} interpolates between flow matching and MeanFlow via a schedule. Our method takes a different approach and share the multi-particle philosophy with IMM \citep{zhou2025inductive} and our focus remains only on the few-step regime (4-50). 

\vspace{-0.1in}
\paragraph{Meta Flow Maps and stochastic flow maps.}
MFM~\citep{potaptchik2026metaflowmapsenable} amortises stochastic flow maps producing i.i.d.\ samples from $p_{1\mid t}(\cdot \mid x_t)$ via diagonal flow matching and consistency losses. MFM and DDM target the same object ($p_{1|t}$) but differ in the training principle --- consistency along auxiliary ODEs versus a proper scoring rule on populations. The emphasis of \citet{potaptchik2026metaflowmapsenable} is reward alignment through efficient value-function estimation from stochastic endpoint samples, whereas our focus is multi-step generation under direct scoring rule training. Our experiments therefore address a different operating regime with late expansion and time-dependent schedules.

\vspace{-0.1em}
\section{Conclusion}

\vspace{-0.1em}
\label{sec:conclusion}

We presented a method that makes efficient stochastic few-step generation with DDMs feasible at DiT scale via two complementary contributions. First, deferred population expansion enables efficient multi-particle training: it shares the early transformer layers across particles and injects particle-specific noise only in a late distributional section, reducing the cost of $m=4$ DDM training from roughly $4\times$ flow matching to about $1.5\times$. Second, time-dependent scoring rule schedules and improved training-time sampling provide time-adaptive distributional objectives, improving few-step performance. 
Overall, our results suggest that shared computation, carefully placed stochasticity, and adaptive objectives make DDMs practical for stochastic few-step generation from scratch.

\vspace{-0.1in}

\paragraph{Limitations.} The regime anchors are derived from a mean-field analysis and are approximate at finite scale. Smooth schedules are more robust than hard steps. At 1-2 steps, distilled and fastforward methods remain stronger. Our gains hold from ImageNet-$256^2$ and transfer to T2I.

\subsection*{AI use statement}

In this work, we used generative AI tools for designing and providing feedback on methodology and experiments, as well as assistance implementing methods.
We have not used generative AI tools for help developing theoretical or conceptual frameworks, formulating mathematical claims, providing critical ingredients for proving mathematical claims, proposing or refining hypotheses, assistance with translation, cleaning and reformatting datasets, support in qualitative and thematic data analysis, and interpretation of results and synthetic data generation, as well as assistance in the writing of proofs, are not applicable to this work. Additionally, we used generative AI tools for creating or modifying scientific figures, creating and editing software code, summarizing existing literature, brainstorming, sourcing/searching for information, editing the research paper for readability, and the identification of relevant literature. We have reviewed all AI-assisted work. Methodology and experiments as well as brainstorming results were discussed among at least two researchers, LLM-generated code was verified and tested for correctness, LLM-assisted literature search was followed up by manual literature review as well as verification of correct summaries by at least two researchers. Finally, AI-generated figures were manually verified and improved. Figures were discussed with at least two researchers to ensure correctness and understandability. We take responsibility for the final content of this work, including text, claims or artifacts produced with the aid of generative AI.

\subsection*{Reproducibility statement}

iDDM is trained using publicly available datasets. We detail data processing and training in Appendix. iDDM is a simple method that works with standard transformers requiring only minimal modifications. Additionally, we detail hyperparameter settings in the Appendix~\ref{app:training_details}. We build upon existing evaluation pipelines and detail evaluation protocols and metrics in Appendix~\ref{app:eval_details}. Therefore, iDDM is easy to reproduce from scratch. To further facilitate reproduction, released code and model checkpoints are available at \url{https://github.com/CompVis/iDDM}.

\bibliography{bibliography}
\bibliographystyle{iclr2027_conference}

\appendix
\clearpage

\section{Model Configuration}
\label{app:training_details}

\newcolumntype{C}{>{\centering\arraybackslash}X}
\begin{table}[htbp]
    \caption{Training configuration for latent-space distributional diffusion on ImageNet $256 \times 256$. DiT-B/2 and DiT-XL/2 use the same recipe except where model scale requires architecture-specific values.}
    \label{tab:training_setup}

    \begin{center}
    \tablestyle{4pt}{1.1}

    \begin{tabularx}{\linewidth}{@{}lCC@{}}
    \toprule
    Config & DiT-B/2 & DiT-XL/2 \\
    \midrule

    \multicolumn{3}{l}{\textit{\textbf{Data and autoencoder}}} \\
    Dataset
        & \multicolumn{2}{c}{ImageNet-256$^2$~\citep{imagenet}} \\
    Preprocessing
        & \multicolumn{2}{c}{resize, center crop, random horizontal flip} \\
    Autoencoder
        & \multicolumn{2}{c}{REPA-E (SD-VAE, f8d4), $32 \times 32 \times 4$~\citep{leng2025repae}} \\

    \midrule
    \multicolumn{3}{l}{\textit{\textbf{Architecture}}} \\
    Depth $L$       & 12   & 28 \\
    Hidden dim      & 768  & 1152 \\
    Head dim        & 64   & 72 \\
    Parameters      & 105M & 515M \\
    Patch size
        & \multicolumn{2}{c}{$2 \times 2$} \\
    Positional encoding
        & \multicolumn{2}{c}{Axial RoPE 2D~\citep{su2024roformer,crowson2024hdit}} \\
    Normalization
        & \multicolumn{2}{c}{Adaptive RMSNorm} \\
    FFN
        & \multicolumn{2}{c}{SwiGLU ($d_{\text{ff}}=3d$)} \\
    Attention scaling
        & \multicolumn{2}{c}{Cosine similarity, learnable per head~\citep{liu2022swin,crowson2024hdit}} \\

    \midrule
    \multicolumn{3}{l}{\textit{\textbf{Conditioning}}} \\
    Class dropout
        & \multicolumn{2}{c}{0.1} \\
    $\xi$ conditioning
        & \multicolumn{2}{c}{Concat.\ fixed Attn./FFN width} \\
    $\xi$ dim
        & \multicolumn{2}{c}{$d_{\text{cat}}=256$ per-token} \\
    $\ell_{\text{start}}$ & 10 & 24 \\

    \midrule
    \multicolumn{3}{l}{\textit{\textbf{Distributional objective}}} \\
    Population size $m$
        & \multicolumn{2}{c}{4} \\
    $\lambda(t)$ \& $\beta(t)$ schedule
        & \multicolumn{2}{c}{linear} \\
    Loss weighting
        & \multicolumn{2}{c}{uniform} \\
    Kernel spatial mode
        & \multicolumn{2}{c}{local} \\

    \midrule
    \multicolumn{3}{l}{\textit{\textbf{Training}}} \\
    Training steps & 400k & 1M \\
    Global batch size
        & \multicolumn{2}{c}{256} \\
    Optimizer
        & \multicolumn{2}{c}{AdamW~\citep{loshchilov2018decoupled}, $(\beta_1,\beta_2)=(0.9,0.95)$} \\
    Dropout
        & \multicolumn{2}{c}{0.0} \\
    Learning rate
        & \multicolumn{2}{c}{$1\times10^{-4}$} \\
    LR schedule
        & \multicolumn{2}{c}{constant, 300-step warmup} \\
    Weight decay
        & \multicolumn{2}{c}{0} \\
    Gradient clipping
        & \multicolumn{2}{c}{1.0} \\
    Precision
        & \multicolumn{2}{c}{bfloat16 AMP} \\
    EMA
        & \multicolumn{2}{c}{EDM-2 EMA~\citep{karras2024analyzing}, $\sigma_{\text{rel}}\in\{0.05,0.10\}$} \\
    $t$ sampler
        & \multicolumn{2}{c}{logitnorm($\mu=-0.8,\sigma=0.8$)} \\

    \midrule
    \multicolumn{3}{l}{\textit{\textbf{Sampling and evaluation}}} \\
    ODE solver
        & \multicolumn{2}{c}{Euler} \\
    steps grid
        & \multicolumn{2}{c}{$\{4,8,16,50\}$} \\

    \bottomrule
    \end{tabularx}
    \end{center}
\end{table}

\clearpage

\begin{table*}[h]
	\centering
	\caption{Training and evaluation configuration for FM and iDDM.
		Block indices are zero-based.}
	\label{tab:t2i_configuration}
	\small
	\setlength{\tabcolsep}{4pt}
	\renewcommand{\arraystretch}{1.1}
	\begin{tabularx}{\linewidth}{@{}lCC@{}}
		\toprule
		Configuration             & FM           & iDDM                              \\
		\midrule
		\multicolumn{3}{l}{\textit{\textbf{Data and architecture}}} \\
		Dataset & \multicolumn{2}{c}{Recaptioned COYO~\citep{kakaobrain2022coyo} subset} \\
		Resolution & \multicolumn{2}{c}{$256\times256$} \\
		Autoencoder & \multicolumn{2}{c}{FLUX.2~\citep{flux2025}, frozen} \\
		Text encoder & \multicolumn{2}{c}{Qwen3-VL-Embedding-2B~\citep{li2026qwen3}, frozen language tower} \\
		Text adapter & \multicolumn{2}{c}{2 layers, output dimension 2048} \\
		Maximum text length & \multicolumn{2}{c}{256 tokens} \\
		Depth / base width & \multicolumn{2}{c}{28 / 1536} \\
		Patch size / head dimension & \multicolumn{2}{c}{$2\times2$ / 96} \\
		\midrule
		\multicolumn{3}{l}{\textit{\textbf{Objective and conditioning}}} \\
		Text dropout & \multicolumn{2}{c}{0.1} \\
		$\xi$ conditioning        & --           & Concat. fixed Attn./FFN width \\
		$\xi$ dim        & --           & $d_\text{cat}=$ 384 per-token \\
		$\ell_\text{start}$        & --           & 20 \\
		Width in blocks 20--25    & 1536         & 1920                              \\
		Population size           & --           & 4                                 \\
		Objective                 & Velocity MSE & Generalized energy-kernel score   \\
		$\lambda(t)$ / $\beta(t)$ & --           & linear, $\beta_\text{start}=0.1$                \\
		Kernel spatial mode       & --           & local                             \\
		Loss weighting & \multicolumn{2}{c}{uniform} \\
		\midrule
		\multicolumn{3}{l}{\textit{\textbf{Training}}} \\
		Evaluated checkpoint & \multicolumn{2}{c}{300k steps} \\
		Global batch size     & \multicolumn{2}{c}{1024} \\
		Optimizer & \multicolumn{2}{c}{AdamW~\citep{loshchilov2018decoupled},
		$(\beta_1,\beta_2)=(0.9,0.95)$} \\
		Dropout & \multicolumn{2}{c}{0.0} \\
		Learning rate & \multicolumn{2}{c}{$1\times10^{-4}$} \\
		LR schedule & \multicolumn{2}{c}{constant, 300-step warmup} \\
		Weight decay & \multicolumn{2}{c}{$1\times10^{-3}$} \\
		Gradient clipping & \multicolumn{2}{c}{1.0} \\
		Precision & \multicolumn{2}{c}{bfloat16 AMP} \\
		Evaluation EMA & \multicolumn{2}{c}{EDM-2 EMA~\citep{karras2024analyzing}, $\sigma_{\text{rel}}\in\{0.05,0.10\}$} \\
		$t$ sampler & \multicolumn{2}{c}{logitnorm($\mu=0,\sigma=1$)} \\
		\midrule
		\multicolumn{3}{l}{\textit{\textbf{Sampling and evaluation}}} \\
		ODE Solver & \multicolumn{2}{c}{Euler} \\
		steps grid & \multicolumn{2}{c}{$\{4,8,16\}$} \\
		CFG search grid & \multicolumn{2}{c}{$\{3,3.5,\ldots,7\}$} \\
		Benchmarks & \multicolumn{2}{c}{MS-COCO FID, CLIPScore, GenEval, T2I-CompBench; HPS v2.1} \\
		\bottomrule
	\end{tabularx}
\end{table*}

\clearpage

\section{Extended Quantitative Evaluations}
\label{app:additional_results}

\subsection{ImageNet $256^2$}
\label{app:eval_details}
All FID scores are computed using 50,000 generated samples compared against the full ImageNet training set statistics, following the ADM evaluation protocol~\citep{dhariwal2021diffusion}. We use the same Inception-V3 features and reference statistics as prior work to ensure comparability. For each model (including methods from other papers), we sweep CFG scales in $\left[1.0, 6.0\right]$ at 0.1 intervals and report the best. Posthoc EMA~\citep{karras2024analyzing} with $\sigma_\text{rel} = 0.05$ is used unless otherwise noted.

For completeness,~\cref{tab:imagenet_256_various_methods} provides broader system-level context on ImageNet-$256^2$ across heterogeneous training paradigms. As they differ in architecture, parameter count, and training data pipeline, they should be read as context rather than a controlled comparison.

\begin{table}[!h]
    \centering
    \caption{\textbf{Reference Methods on ImageNet-256$^\mathbf{2}$.} FID$\downarrow$ of representative methods from other model families, as reported in the respective publications.}
    \label{tab:imagenet_256_various_methods}
    \newcommand{\group}[1]{\addlinespace[.5em]\multicolumn{4}{@{}l}{\textit{\textbf{#1}}}\\\addlinespace[.15em]}
    \newcommand{\firstgroup}[1]{\multicolumn{4}{@{}l}{\textit{\textbf{#1}}}\\\addlinespace[.15em]}
    \scalebox{0.8}{%
        \begin{tabular}{>{\hspace{.6em}}lccc}
            \toprule
            \multicolumn{1}{@{}l}{Method} & Parameters & Steps & FID$\downarrow$ \\
            \midrule
            \firstgroup{GANs}
            StyleGAN-XL~\citep{sauer2022stylegan} & 166M & 1 & 2.30 \\
            \group{Autoregressive/Masking}
            STARFlow~\citep{gu2025starflow} & 1.4B & -- & 3.80 \\
            VAR-$d$30~\citep{tian2024visual} & 2B & 10 & 1.92 \\
            \group{Many-step diffusion/FM}
            ADM-G~\citep{dhariwal2021diffusion} & 554M & 250 & 4.59 \\
            DiT-XL/2~\citep{peebles2023dit} & 675M & 250 & 2.27 \\
            SiT-XL/2~\citep{ma2024sit} & 675M & 250 & 2.06 \\
            LightningDiT-XL/2~\citep{yao2025vavae} & 675M & 250 & 1.35 \\
            RAE + $\text{DiT}\smash{^{\text{DH}}}\text{-XL}$~\citep{zheng2025diffusion} & 839M & 50 & 1.13 \\
            \group{Consistency Models}
            iCT-XL/2~\citep{song2024improved} via~\citep{zhou2025inductive} & 675M & 2 & \llap{2}0.3\phantom{0} \\
            FACM~\citep{peng2026facm} & 675M & 2 & 1.32 \\
            \group{Flow Maps}
            TVM-XL/2~\citep{tvm2025} & 678M & 4 & 1.99 \\
            $\alpha\text{-Flow-XL/2+}$~\citep{zhang2026alphaflow} & 676M & 2 & 1.95 \\
            DMF-XL/2~\citep{lee2026decoupled} & 675M & 4 & 1.51 \\
            \group{Direct Distribution Matching}
            Drifting Model, L/2~\citep{deng2026drifting} & 463M & 1 & 1.54 \\
            W-Flow, XL/2~\citep{han2026onestepgenerativemodelingwasserstein} & 679M & 1 & 1.29  \\
            TBSM-XL/2~\citep{sun2026threebodyscatteringgenerativemodeling} & 675M & 1 & 1.63 \\
            \bottomrule
        \end{tabular}%
    }
\end{table}

\paragraph{Additional Metrics.}
In addition to FID, we report FD-DINOv2, precision, and recall. FD-DINOv2 follows Stein et al.~\citep{stein2023exposing} and computes the Fr\'echet distance in DINOv2-ViT-L/14 feature space. Precision and recall follow Kynk\"a\"anniemi et al.~\citep{kynkaanniemi2019improved}, using $k$-nearest-neighbor estimates of the real and generated manifolds. Precision reflects sample fidelity, while recall reflects distributional coverage, with higher values being better for both. KID follows Bi\'nkowski et al.~\citep{binkowski2018demystifying} and computes the squared maximum mean discrepancy between Inception-V3 features of real and generated images using a polynomial kernel. KDD computes the analogous kernel distance in DINOv2-ViT-L/14 feature space. Inception Score follows Salimans et al.~\citep{salimans2016improved} and measures the KL divergence between the conditional class distribution predicted by an Inception-V3 classifier and its marginal over the generated set. Precision and recall follow Kynk\"a\"anniemi et al.~\citep{kynkaanniemi2019improved}, using $k$-nearest-neighbor estimates of the real and generated manifolds. Precision reflects sample fidelity, while recall reflects distributional coverage, with higher values being better for both.

\paragraph{Additional Ablations.}
To verify that the gains in \cref{tab:main_ablation} are not an artifact of the A$\to$F ordering, \cref{tab:independent_component_ablation} adds each refinement independently to \textsc{Config B} rather than cumulatively. All three components improve 4-step FID in isolation (17.51 $\to$ 17.21 for $\xi$-conditioning, $\to$17.34 for $t$-sampling, $\to$15.77 for the $(\lambda,\beta)$ schedule), confirming that none of the gains in \cref{tab:main_ablation} depends on the presence of the other components. Improved $\xi$-conditioning is the most uniformly beneficial change, improving FID, FD-DINOv2, KID, KDD, and precision at both step counts, at the cost of a small reduction in 4-step recall (0.154 $\to$ 0.139).

\cref{tab:all_metrics} extends the main FM/DDM/iDDM comparison of \cref{tab:main_ablation} with the same additional metrics. At 4 steps, iDDM improves over every metric relative to both baselines simultaneously: FID (43.6$\to$13.2 vs. DDM), FD-DINOv2 (801.8$\to$318.8), KID (0.0337$\to$0.0030), KDD (1.854$\to$0.335), Inception Score (40.3$\to$162.5), precision (0.704$\to$0.778), and recall (0.164$\to$0.226) -- ruling out a fidelity/diversity trade-off as the source of iDDM's FID gain over naïve DDM. iDDM also matches or exceeds the FM reference on every metric at both 4 and 50 steps at matched training steps, including precision and recall, which are otherwise close to identical between the two at 50 steps (0.834 vs. 0.833 precision, 0.592 vs. 0.590 recall). Naïve DDM's precision remains close to FM's at both step counts (and is marginally higher at 50 NFE), indicating that its poor FID, FD-DINOv2, and Inception Score at 4 steps reflect a broader degradation in sample quality rather than a precision-specific failure.

\begin{table}[!h]
\caption{\textbf{Independent Component Ablations.} We independently add each proposed component to \textsc{Config B} and evaluate generation quality across multiple metrics.}
\label{tab:independent_component_ablation}
\begin{center}
\setlength{\tabcolsep}{4pt}
\resizebox{0.95\linewidth}{!}{
\begin{tabular}{clrrrrrrr}
\toprule
NFE
& Configuration
& FID $\downarrow$
& FD-DINOv2 $\downarrow$
& KID $\downarrow$
& KDD $\downarrow$
& IS $\uparrow$
& Prec. $\uparrow$
& Rec. $\uparrow$ \\
\midrule

\multirow{4}{*}{4}
& \textsc{Config} \textbf{B}
& 17.510 & 496.437 & 0.00416 & 0.7158 & 128.48 & 0.757 & 0.154 \\
& \textbf{B} + Improved $\xi$-conditioning
& 17.214 & 464.758 & 0.00403 & 0.6272 & 128.85 & \textbf{0.773} & 0.139 \\
& \textbf{B} + Improved $t$-sampling
& 17.338 & 528.844 & 0.00418 & 0.8124 & \textbf{134.44} & 0.763 & 0.153 \\
& \textbf{B} + $(\lambda,\beta)$ schedule
& \textbf{15.770} & \textbf{353.751} & \textbf{0.00384} & \textbf{0.3968} & 130.25 & 0.738 & \textbf{0.226} \\
\midrule

\multirow{4}{*}{50}
& \textsc{Config} \textbf{B}
& 5.447 & 204.023 & 0.00130 & 0.2354 & 142.84 & 0.840 & 0.568 \\
& \textbf{B} + Improved $\xi$-conditioning
& 5.356 & 195.764 & 0.00123 & 0.2211 & 145.51 & \textbf{0.847} & 0.563 \\
& \textbf{B} + Improved $t$-sampling
& \textbf{5.141} & 225.909 & \textbf{0.00108} & 0.2813 & \textbf{161.42} & 0.836 & \textbf{0.576} \\
& \textbf{B} + $(\lambda,\beta)$ schedule
& 5.851 & \textbf{168.054} & 0.00158 & \textbf{0.1660} & 143.15 & 0.837 & 0.562 \\

\bottomrule
\end{tabular}}
\end{center}
\end{table}

\begin{table}[!h]
\caption{\textbf{Additional Evaluation Metrics.} We add additional metrics commonly used for evaluation on ImageNet-256 for our final models.}
\label{tab:all_metrics}
\begin{center}
\resizebox{0.8\linewidth}{!}{
\begin{tabular}{clrrrrrrr}
\toprule
NFE & Method & FID $\downarrow$ & FD-DINOv2 $\downarrow$ & KID $\downarrow$ &
KDD $\downarrow$ & IS $\uparrow$ & Prec. $\uparrow$ & Rec. $\uparrow$ \\
\midrule

\multirow{3}{*}{4}
& FM   & 26.699 & 418.711 & 0.00870 & 0.5153 & 99.10  & 0.712 & 0.116 \\
& DDM  & 43.637 & 801.812 & 0.03367 & 1.8544 & 40.33  & 0.704 & 0.164 \\
& iDDM & \textbf{13.224} & \textbf{318.794} & \textbf{0.00300} & \textbf{0.3348} & \textbf{162.45} & \textbf{0.778} & \textbf{0.226} \\
\midrule

\multirow{3}{*}{50}
& FM   & 4.970  & 175.700 & 0.00121 & 0.1869 & 151.30 & 0.833 & 0.590 \\
& DDM  & 10.094 & 244.954 & 0.00300 & 0.2768 & 111.77 & \textbf{0.845} & 0.487 \\
& iDDM & \textbf{4.614}  & \textbf{173.983} & \textbf{0.00109} & \textbf{0.1833} & \textbf{159.43} & 0.834 & \textbf{0.592} \\

\bottomrule
\end{tabular}}
\end{center}
\end{table}

\paragraph{Matched Compute Comparison with Flow Matching} We report iDDM at fractions of its full training run, with the FM baseline extended correspondingly to 600k train steps, so both are compared at matched compute budget rather than matched update count. Training compute is normalized to FM's full budget (400k steps); per-step training cost is 1× (FM), 1.5× (iDDM), 4× (DDM). Step counts are chosen so FM and iDDM are compute-matched exactly (up to checkpoint granularity of 10k steps). Where exact matching is off-grid, iDDM and DDM checkpoints are \emph{rounded down} to the nearest 10k steps, \emph{i.e.}, they receive at most equal training compute. $\dagger$ marks each method's full training run as reported in the paper. The CFG scale is selected independently per method and per inference-step budget with respect to FID; FD-DINOv2 is reported at that same CFG. All other evaluation settings (number of samples, reference statistics, sampler) are identical across all cells.

\begin{table}[!h]
\caption{Compute-matched comparison for 4-step and 50-step inference. $\dagger$ denotes the reference compute settings.}
\label{tab:compute_scaling}
\begin{center}
\small

\textbf{(a) 4-step inference}\\[3pt]

\begin{tabular*}{\linewidth}{@{\extracolsep{\fill}}lrrrrrrrrr@{}}
\toprule
& \multicolumn{3}{c}{Training Steps}
& \multicolumn{3}{c}{FID $\downarrow$}
& \multicolumn{3}{c}{FD-DINOv2 $\downarrow$} \\
\cmidrule(lr){2-4}
\cmidrule(lr){5-7}
\cmidrule(lr){8-10}
Compute
& FM & iDDM & DDM
& FM & iDDM & DDM
& FM & iDDM & DDM \\
\midrule
$0.60\times$          & 240k & 160k & 60k  & 29.04 & 16.94 & 67.85 & 462.63 & 416.87 & 1091.38 \\
$0.75\times$          & 300k & 200k & 70k  & 27.07 & 15.80 & 64.01 & 437.26 & 382.58 & 1060.85 \\
$0.90\times$          & 360k & 240k & 90k  & 27.85 & 14.43 & 63.05 & 434.98 & 362.17 & 1021.61 \\
$1.00\times^\dagger$  & 400k & 260k & 100k & 26.70 & 14.33 & 59.91 & 418.71 & 355.91 & 999.49 \\
$1.20\times$          & 480k & 320k & 120k & 27.70 & 13.69 & 53.09 & 436.41 & 337.14 & 927.10 \\
$1.35\times$          & 540k & 360k & 130k & 27.00 & 13.25 & 55.47 & 413.03 & 326.29 & 945.77 \\
$1.50\times^\dagger$  & 600k & 400k & 150k & 26.36 & 13.13 & 52.63 & 406.10 & 318.79 & 915.72 \\
\bottomrule
\end{tabular*}

\vspace{6pt}

\textbf{(b) 50-step inference}\\[3pt]

\begin{tabular*}{\linewidth}{@{\extracolsep{\fill}}lrrrrrrrrr@{}}
\toprule
& \multicolumn{3}{c}{Training Steps}
& \multicolumn{3}{c}{FID $\downarrow$}
& \multicolumn{3}{c}{FD-DINOv2 $\downarrow$} \\
\cmidrule(lr){2-4}
\cmidrule(lr){5-7}
\cmidrule(lr){8-10}
Compute
& FM & iDDM & DDM
& FM & iDDM & DDM
& FM & iDDM & DDM \\
\midrule
$0.60\times$          & 240k & 160k & 60k  & 6.35 & 7.13 & 32.92 & 195.51 & 223.20 & 535.76 \\
$0.75\times$          & 300k & 200k & 70k  & 5.64 & 6.42 & 30.25 & 186.62 & 208.76 & 492.68 \\
$0.90\times$          & 360k & 240k & 90k  & 5.32 & 5.79 & 26.03 & 171.86 & 202.33 & 438.58 \\
$1.00\times^\dagger$  & 400k & 260k & 100k & 4.97 & 5.62 & 23.60 & 175.70 & 182.12 & 421.34 \\
$1.20\times$          & 480k & 320k & 120k & 4.63 & 5.07 & 20.74 & 177.23 & 176.69 & 371.56 \\
$1.35\times$          & 540k & 360k & 130k & 4.41 & 4.81 & 19.35 & 169.73 & 181.21 & 362.33 \\
$1.50\times^\dagger$  & 600k & 400k & 150k & 4.21 & 4.61 & 17.76 & 163.87 & 173.98 & 331.29 \\
\bottomrule
\end{tabular*}

\end{center}
\end{table}

\paragraph{Posterior Diversity}
\label{app:posterior-div}

Direct posterior metrics (\emph{e.g.}, MMD or mean and variance errors against $p(x_1\mid x_t)$) require reference samples from the posterior, which are unavailable at this scale, and importance-weighting the training set degenerates in the 4096-dimensional latent space. We therefore measure how much a model's samples vary for a fixed $x_t$, and whether that variation depends on $x_t$. For each of the 1,000 ImageNet classes, we noise one training image to a fixed $t$, draw 50 samples from each resulting $x_t$ with 8 Euler steps from $t$ to 1, and compute the FID of the 50,000 pooled samples against the standard full-dataset reference statistics (DiT-B;~\cref{tab:cond-div}).

A deterministic predictor produces only 1,000 distinct images, so its FID stays at this limit for every $t$, FM indeed shows this with FID$\approx 38$ independent of $t$. A predictor that ignored $x_t$ would also be flat in $t$, but low. A sampler that follows the posterior must instead be diverse at low $t$ and approach the 1,000-image limit as the posterior concentrates, so its FID increases with $t$. Both distributional models show this profile. iDDM is more diverse than DDM at low $t$ and less at high $t$, consistent with its schedule, which moves the objective toward regression as the posterior concentrates. This diagnostic detects missing or unconditioned diversity, it does not measure posterior accuracy.

\begin{table}[h]
\centering\small
\caption{FID@50k with samples drawn starting from 1{,}000 fixed $x_t$ (DiT-B, 8 steps).}
\label{tab:cond-div}
\begin{tabular}{lcccc}
\toprule
Method & $t{=}0.2$ & $t{=}0.4$ & $t{=}0.6$ & $t{=}0.8$ \\
\midrule
FM (deterministic) & 37.90 & 37.70 & 37.81 & 38.75 \\
DDM   & 17.31 & 16.54 & 17.64 & 21.07 \\
iDDM  & 13.77 & 16.12 & 18.96 & 23.88 \\
\bottomrule
\end{tabular}
\end{table}

\paragraph{Training Compute Comparison} To fairly compare training compute, we reimplemented iDDM in JAX inside the MeanFlow/iMF repository and ran a performance benchmark on $4\times$A100-80GB with local batch 32 / global 128 and DiT-XL/2 architecture. As can be seen in~\cref{tab:training_efficiency}, iDDM is the cheapest of the three alternatives per step. Combined with the reported training lengths (240 epochs for MeanFlow-XL/2 and 800 for iMF-XL/2 against 200 for iDDM-XL/2) MeanFlow's total training budget is roughly $1.67\times$ ours and iMF's roughly $17.6\times$ ours.

\begin{table}[!h]
\caption{Training throughput benchmarked on $4\times$A100-80GB, local batch 32 / global 128, DiT-XL/2}
\label{tab:training_efficiency}
\begin{center}
\begin{tabular}{lrrrrr}
\toprule
Method & it/s & img/s & ms/it & $\times$ slower vs.\ FM \\
\midrule
FM   & 2.79 & 357 & 358.1  & 1.00$\times$ \\
iDDM & 1.98 & 254 & 504.2  & 1.41$\times$ \\
MeanFlow   & 1.42 & 182 & 705.1  & 1.96$\times$ \\
iMF  & 0.45 & 57  & 2228.2 & 6.20$\times$ \\
\bottomrule
\end{tabular}
\end{center}
\end{table}

\subsection{Text-to-Image}
\label{app:text_to_image}

\paragraph{Architecture.}
Our backbone consists of 28 transformer blocks with hidden dimension 1536 and head dimension 96. We use adaptive RMSNorm~\citep{zhang2019root}, SwiGLU feedforward layers~\citep{shazeer2020glu} with expansion factor 3, and axial rotary position embeddings~\citep{su2024roformer,crowson2024hdit}. Attention uses cosine similarity with learnable per-head scaling~\citep{liu2022swin}.

\paragraph{Model.}
We encode images using the frozen FLUX.2 autoencoder~\citep{flux2025}, producing $32\times32\times32$ latents at $256\times256$ resolution, and use $2\times2$ latent patches. Text conditioning uses the frozen language tower of Qwen3-VL-Embedding-2B~\citep{li2026qwen3}, followed by two trainable transformer layers. FM and iDDM share this backbone. For iDDM, we concatenate 384 gated noise channels before block 20 and remove them before block 26, increasing the width of these six blocks to 1920. Block indices are zero-based.

\paragraph{Data and Training.} 
Both models use a recaptioned subset of COYO~\citep{kakaobrain2022coyo}. FM minimizes velocity prediction error, while iDDM uses a generalized energy-kernel objective with four predictions per training example. We evaluate the 300k-step checkpoints using post-hoc EMA~\citep{karras2024analyzing}. Table~\ref{tab:t2i_configuration} summarizes the configuration.

\paragraph{Evaluation.}
Fidelity to natural images is measured by COCO-FID, computed against 30{,}000 reference images following the standard MS-COCO evaluation protocol, together with CLIP Cosine similarity between generated images and their conditioning captions. We additionally evaluate compositional alignment on GenEval~\citep{ghosh2023geneval} and T2I-CompBench~\citep{huang2023t2icompbench}, and human preference alignment using HPS v2.1~\citep{wu2023hpsv2}. We use uniform Euler sampling with 4, 8, and 16 steps and sweep CFG for each model and step combination for FID and CLIP Cosine similarity jointly, and sweep individually for HPS v2.1 and T2I-CompBench.

\begin{table}[!h]
	\caption{Performance at 16, 8, and 4 sampling steps. Parentheses show absolute changes relative to each method's 16-step result. For HPS v2.1 and T2I-CompBench, CFG is selected separately per method, step count, and benchmark. T2I-CompBench selection maximizes the mean of its six category scores.}
	\begin{center}
		\begin{tabular}{llrrr}
			\toprule
			Metric & Method & 16 steps (reference) & 8 steps ($\Delta$) & 4 steps ($\Delta$) \\
			\midrule
			
			\multirow{2}{*}{\textbf{FID $\downarrow$}}
			       & FM     & 11.200               & 26.327 (+15.127)   & 78.195 (+66.995)   \\
			       & iDDM   & 11.265               & 19.403 (+8.138)    & 41.046 (+29.781)   \\
			\midrule
			
			\multirow{2}{*}{\textbf{CLIP Cosine $\uparrow$}}
			       & FM     & 0.3141               & 0.3080 ($-0.0061$) & 0.2744 ($-0.0397$) \\
			       & iDDM   & 0.3102               & 0.3030 ($-0.0072$) & 0.2840 ($-0.0262$) \\
			\midrule
			
			\multirow{2}{*}{\textbf{GenEval $\uparrow$}}
			       & FM     & 0.364                & 0.285 ($-0.079$)   & 0.122 ($-0.242$)   \\
			       & iDDM   & 0.353                & 0.297 ($-0.056$)   & 0.209 ($-0.144$)   \\
			\midrule
			
			\multirow{2}{*}{\textbf{HPS v2.1 $\uparrow$}}
			       & FM     & 22.845               & 19.141 ($-3.703$)  & 13.913 ($-8.932$)  \\
			       & iDDM   & 22.510               & 19.820 ($-2.691$)  & 16.968 ($-5.542$)  \\
			\midrule
			
			\multicolumn{5}{l}{\textbf{T2I CompBench}} \\[1pt]
			
			\multirow{2}{*}{\hspace{1em}Color $\uparrow$}
			       & FM     & 0.6786               & 0.5282 ($-0.1505$) & 0.3185 ($-0.3602$) \\
			       & iDDM   & 0.5680               & 0.4820 ($-0.0860$) & 0.4151 ($-0.1529$) \\
			\cmidrule(lr){1-5}
			
			\multirow{2}{*}{\hspace{1em}Shape $\uparrow$}
			       & FM     & 0.4351               & 0.3476 ($-0.0875$) & 0.2484 ($-0.1867$) \\
			       & iDDM   & 0.3857               & 0.3406 ($-0.0451$) & 0.3061 ($-0.0796$) \\
			\cmidrule(lr){1-5}
			
			\multirow{2}{*}{\hspace{1em}Texture $\uparrow$}
			       & FM     & 0.6274               & 0.4960 ($-0.1314$) & 0.3189 ($-0.3085$) \\
			       & iDDM   & 0.5932               & 0.5348 ($-0.0585$) & 0.4755 ($-0.1177$) \\
			\cmidrule(lr){1-5}
			
			\multirow{2}{*}{\hspace{1em}Spatial $\uparrow$}
			       & FM     & 0.2901               & 0.1628 ($-0.1272$) & 0.0443 ($-0.2457$) \\
			       & iDDM   & 0.1178               & 0.0657 ($-0.0521$) & 0.0232 ($-0.0946$) \\
			\cmidrule(lr){1-5}
			
			\multirow{2}{*}{\hspace{1em}Non-spatial $\uparrow$}
			       & FM     & 0.3021               & 0.2879 ($-0.0142$) & 0.2532 ($-0.0489$) \\
			       & iDDM   & 0.2923               & 0.2809 ($-0.0114$) & 0.2596 ($-0.0327$) \\
			\cmidrule(lr){1-5}
			
			\multirow{2}{*}{\hspace{1em}Complex $\uparrow$}
			       & FM     & 0.3517               & 0.2956 ($-0.0561$) & 0.2102 ($-0.1415$) \\
			       & iDDM   & 0.3525               & 0.3234 ($-0.0291$) & 0.2891 ($-0.0634$) \\
			
			\bottomrule
		\end{tabular}
	\end{center}
\end{table}

\section{Dynamical Regimes of Diffusion Models}
\label{app:dynamical_regimes}

This appendix gives the details needed for the schedule construction in Section~\ref{sec:dynamical_regimes}. The central object is not an equivalence between the OU reverse SDE and the flow-matching probability path, but the conditional law induced by an isotropic Gaussian corruption after normalization.

\subsection{SNR coordinate and OU/FM mapping}
\label{app:forward_processes}

Consider any isotropic Gaussian corruption
\begin{equation}\label{eq:snr_definitions}
    x = \alpha x_1 + \sigma \epsilon,
    \qquad
    \epsilon \sim \mathcal{N}(0,I),
    \qquad
    \rho = \frac{\alpha^2}{\sigma^2}.
\end{equation}
For $\alpha > 0$, the deterministic normalization
\begin{equation}\label{eq:normalized_denoising}
    z = \frac{x}{\alpha} = x_1 + \rho^{-1/2}\epsilon
\end{equation}
shows that the conditional law $p(x_1\mid z)$ depends on the corruption process through the normalized observation $z$ and the scalar $\rho$. This is the only sense in which we compare OU and FM time: matched SNR defines the same normalized denoising problem, not the same stochastic process, reverse dynamics, path measure, or raw-coordinate marginal entropy.

The Ornstein--Uhlenbeck forward process used by \citet{Biroli2024}, which is the constant-rate case of the VP-SDE family, is
\begin{equation}\label{eq:ou_forward}
    \rmd x_\tau = -x_\tau\,\rmd \tau + \sqrt{2}\,\rmd W_\tau, \qquad x_{\tau=0} \sim p_\text{data}(x)\,,
\end{equation}
with solution kernel
\begin{equation}\label{eq:ou_kernel}
    x_\tau \mid x_1 \sim \mathcal{N}\!\left(e^{-\tau} x_1,\; (1 - e^{-2\tau})\, I\right)\,, \qquad \mathrm{SNR}_\text{OU}(\tau) = \frac{e^{-2\tau}}{1 - e^{-2\tau}}\,.
\end{equation}
Our models use the standard linear interpolant $x_t = (1-t)x_0 + t\,x_1$, $x_0 \sim \mathcal{N}(0,I)$, $t \in [0,1]$. The corresponding kernel is
\begin{equation}\label{eq:fm_kernel}
    x_t \mid x_1 \sim \mathcal{N}\!\left(t\,x_1,\; (1-t)^2 I\right)\,, \qquad \mathrm{SNR}_\text{FM}(t) = \frac{t^2}{(1-t)^2}\,.
\end{equation}
Equating the two SNRs gives the coordinate conversion
\begin{equation}\label{eq:snr_bijection}
    t = \frac{\sqrt{\rho}}{1 + \sqrt{\rho}} = \frac{e^{-\tau}}{e^{-\tau} + \sqrt{1 - e^{-2\tau}}},
    \qquad
    \tau = -\tfrac{1}{2}\log\!\frac{\rho}{1 + \rho}.
\end{equation}
Applying Eq.~\ref{eq:snr_bijection} to a threshold means only that the two corruptions induce the same normalized denoising problem of Eq.~\ref{eq:normalized_denoising}.

\subsection{Regime anchors}
\label{app:regime_thresholds}

\paragraph{Speciation $\rho_s$.}
\citet{Biroli2024} define $\tau_s$ as the OU time at which the noise variance along the principal axis matches the signal variance, $\Lambda \cdot e^{-2\tau_s} = 1$. Equivalently, the variance of the leading mode of the OU noised marginal becomes informative about the data-manifold direction at this point. Expressed as an SNR value, $e^{-2\tau_s} = 1/\Lambda$ gives
\begin{equation}
    \rho_s = \frac{1/\Lambda}{1 - 1/\Lambda} = \frac{1}{\Lambda - 1}.
\end{equation}
For $\Lambda \gg 1$, $\rho_s \approx 1/\Lambda$. We use this OU-derived SNR as a schedule anchor; its FM coordinate is $t_s^\text{FM} = 1/(1 + \sqrt{\Lambda - 1})$.

\paragraph{Collapse $\rho_c$.}
The collapse criterion of \citet{Biroli2024} is OU-specific: it balances the differential entropy density of the empirical OU noised marginal $H(\tau)$ against the per-coordinate entropy density of $n$ well-separated Gaussian blobs of variance $\Delta_\tau = 1 - e^{-2\tau}$,
\begin{equation}\label{eq:collapse}
    H(\tau_c) = H^\text{sep}(\tau_c),
    \qquad
    H^\text{sep}(\tau) = \frac{\log n}{d} + \tfrac{1}{2}\bigl(1 + \log(2\pi \Delta_\tau)\bigr).
\end{equation}
The collapse time $\tau_c$ is the smallest OU time at which Eq.~\ref{eq:collapse} holds; below $\tau_c$, the empirical OU noised distribution is geometrically dominated by isolated training-example components, and a typical normalized observation has an identity posterior concentrated on a single index. We convert this OU diagnostic to an SNR anchor by setting
\begin{equation}
    \rho_c = \frac{e^{-2\tau_c}}{1 - e^{-2\tau_c}}.
\end{equation}
Because $H(\tau)$ has no closed form, $\tau_c$ is computed numerically following \citet{Biroli2024}.

\paragraph{Separation $\rho_\text{sep}(\varepsilon)$.}
For the empirical finite-sample distribution, we use a geometric containment criterion: the $(1-\varepsilon)$-quantile OU noise ball around a training point should fit inside half the nearest-neighbor distance. With
\[
    d_\text{min} = \min_{i\neq j}\|x_1^{(i)}-x_1^{(j)}\|,
\]
and with $\chi^2_{d,1-\varepsilon}$ denoting the $(1-\varepsilon)$-quantile of a $\chi^2_d$ random variable, this criterion is
\begin{equation}
    \sqrt{1-e^{-2\tau_{\text{sep}}}}\,\sqrt{\chi^2_{d,1-\varepsilon}}
    =
    e^{-\tau_{\text{sep}}}\,\frac{d_{\mathrm{min}}}{2}.
\end{equation}
Solving and converting to SNR gives
\begin{equation}
\label{eq:rho_sep}
    \rho_\text{sep}(\varepsilon)
    =
    \frac{e^{-2\tau_\text{sep}}}{1-e^{-2\tau_\text{sep}}}
    =
    \frac{4\chi^2_{d,1-\varepsilon}}{d_\text{min}^2}.
\end{equation}
This is a high-probability containment anchor. As with $\rho_c$, its ordering relative to the other anchors should be treated as a measured property of the dataset rather than a universal theorem.

\subsection{ImageNet latent estimates and schedule construction}
\label{app:rho_numerical}

We compute the characteristic times on REPA-E~\citep{leng2025repae} (SD-VAE backbone) latents of ImageNet-256. Images are center-cropped to $256{\times}256$, rescaled to $[-1,1]$, and encoded with the frozen autoencoder; the resulting latent is $32{\times}32{\times}4$, of which we keep the first channel only, giving feature dimension $d = 32{\times}32 = 1024$. The analyzed set $\{x_1^{(i)}\}_{i=1}^n$ is a uniformly sampled subset of $n = 50{,}000$ training latents drawn across all 1{,}000 ImageNet classes, centered to zero per-coordinate mean. From this set: $\Lambda$ is the largest eigenvalue of the empirical $d \times d$ covariance, giving $\rho_s$; $\rho_c$ is obtained by solving the OU entropy balance $H(\tau_c) = H^\text{sep}(\tau_c)$ numerically, with the empirical entropy density estimated by Monte Carlo from the $n$-component mixture using $n' = 20{,}000$ noisy probe samples; $d_\text{min}$ is computed by an exact pairwise distance scan over the centered set, and used with containment level $1-\varepsilon = 0.9$ to instantiate $\rho_\text{sep}(\varepsilon)$ via Eq.~\ref{eq:rho_sep}. Table~\ref{tab:characteristic-times} reports the resulting SNR characteristic times and their OU/FM coordinate images.

\begin{table}[t]
\centering
\caption{Characteristic times on centered first-channel REPA-E latents for ImageNet-256 ($n=50{,}000$ samples drawn across all classes). $\rho_s$ and $\rho_c$ are OU-derived posterior-concentration diagnostics; $\rho_{\mathrm{sep}}$ is an empirical high-probability containment diagnostic using the $\chi^2$ noise norm. The $\tau$ and $t$ columns are coordinate images from Eq.~14, not an equivalence of OU and FM dynamics.}
\label{tab:characteristic-times}
\begin{tabular}{llccc}
\toprule
Threshold & Definition & $\rho$ & $\tau_{\mathrm{OU}}$ & $t_{\mathrm{FM}}$ \\
\midrule
$\rho_s$                   & $1/(\Lambda-1)$                      & 0.015 & 2.11  & 0.11 \\
$\rho_c$                   & $H(\tau_c)=H^{\mathrm{sep}}(\tau_c)$ & 0.03  & 1.77  & 0.15 \\
$\rho_{\mathrm{sep}}(0.1)$ & $4\chi^2_{d,0.9}/d_{\min}^2$         & 33.11 & 0.015 & 0.85 \\
\bottomrule
\end{tabular}
\end{table}

We instantiate our piecewise schedule to be 1 when $\rho<\rho_s$, 0 when $\rho > \rho_\text{sep}$ and a linear interpolation in between.

\section{On Scheduling $\lambda(t)$ and $\beta(t)$}
\label{app:on_lambda_beta}

 Regarding \((\lambda(t),\beta(t))\) schedules, conditional on introducing regression bias, its orientation toward the high-SNR/data end minimizes an explicit posterior-covariance bias criterion.  

In our FM convention,
$$ Y_\rho=X_1+\rho^{-1/2}Z,\qquad \rho=\frac{t^2}{(1-t)^2}. $$
For $X_1\in L^2$, the average posterior variance
$$v(\rho)=d^{-1}\mathbb E\,\mathrm{tr}\,\mathrm{Cov}(X_1\mid Y_\rho) $$
satisfies $v(\rho)\le\rho^{-1}$, since the conditional mean has no larger MSE than the estimator $Y_\rho$. It is also non-increasing in $\rho$, because a lower-SNR Gaussian observation is a degradation of a higher-SNR one. Thus, throughout IIIb, $v(\rho)\le\rho_{\mathrm{sep}}^{-1}$.

Under the isotropic-Gaussian population surrogate of~\citet{debortoli2025distributional}, the generalized energy-score optimum preserves the conditional mean and multiplies its covariance by
$$ f(\lambda,\beta)=\left(2\lambda^{-2/(2-\beta)}-1\right)^{-1}, $$
for $0<\lambda\le1$, $0<\beta<2$. Hence, writing $a_t=1-f_t$, the expected covariance removed per coordinate is
$$ a_t v(\rho_t)\le \frac{a_t}{\rho_t}. $$
A finite FM transition introduces only the additional factor $((u-t)/(1-t))^2$; Proposition C.7 in~\citet{debortoli2025distributional} gives the analogous decomposition for the Gaussian diffusion bridge.

Finally, for a fixed collection of shrinkage values, define
$$ \mathcal B(a)=\int a_t v(\rho_t)\,d\nu(t). $$
If $t_1<t_2$, then $v(\rho_{t_1})\ge v(\rho_{t_2})$; swapping incorrectly ordered values $a_{t_1}>a_{t_2}$ changes $\mathcal B$ by
$$ (a_{t_1}-a_{t_2}) \bigl(v(\rho_{t_2})-v(\rho_{t_1})\bigr)\le0. $$

Therefore $\mathcal B$ is minimized by assigning the largest shrinkage to the highest SNRs. This proves the schedule orientation for this covariance criterion, not a uniquely optimal smooth profile or terminal-FID optimum. The regime anchors motivate the transition locations, while the precise profile and separate $\lambda/\beta$ schedules remain empirical and are ablated in~\cref{tab:train_objective_params_ablation}.

Empirically, we consider the following shape functions:
\begin{align}
    \text{Constant:}\quad
        & s(t) = 1
        \quad\big(\lambda(t)=\lambda_{\max},\ \beta(t)=\beta_{\min}\big), \label{eq:sched_const}\\[2pt]
    \text{Linear:}\quad
        & s(t) = 1-t, \label{eq:sched_linear}\\[2pt]
    \text{Dynamical regimes:}\quad
        & s(t) = \operatorname{clip}\!\left(\frac{t_{\mathrm{sep}}-t}{t_{\mathrm{sep}}-t_s},\,0,\,1\right), \label{eq:sched_dynreg}\\[2pt]
    \text{Step:}\quad
        & s(t) = \mathbb{1}\!\left[t \le \kappa\right], \label{eq:sched_step}\\[2pt]
    \text{SNR:}\quad
        & s(t) = \frac{(1-t)^{2p}}{(1-t)^{2p}+t^{2p}}
        = \frac{1}{1+\operatorname{SNR}(t)^{p}},
        \qquad \operatorname{SNR}(t)=\frac{t^2}{(1-t)^2}, \label{eq:sched_snr}
\end{align}
where step-$\kappa$ in~\cref{tab:train_objective_params_ablation} refers to Eq.~\ref{eq:sched_step}.

\begin{figure}
    \centering
    \includegraphics[width=.5\linewidth]{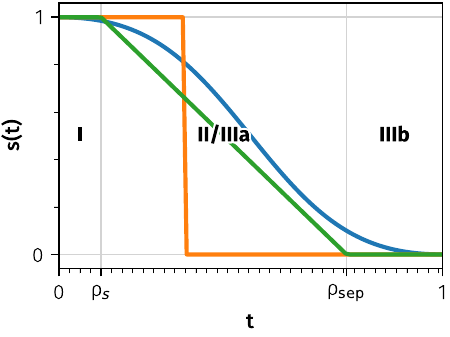}
    \caption{Schedule \& Regime Illustration.}
    \label{fig:schedule_illustration}
\end{figure}

Finally, we provide some visualization of the different schedules as a function of the SNR in Figure \ref{fig:schedule_illustration}.

\section{$\xi$ Conditioning}
\label{app:xi_ablation}
We explore various conditioning mechanisms, summarized in \cref{sec:efficient_training} and \cref{fig:xi_conditioning,tab:xi_conditioning_hparam_ablation}.
After deferred expansion, each population member carries an independent noise tensor $\xi \in \mathbb{R}^{H'\times W'\times d_\mathrm{noise}}$ that must be injected into the model (starting) at layer $\ell_\text{start}$.
We consider four relevant families of mechanisms, each combined with adaptive gating, detailed in the following paragraphs.

\paragraph{Adaptive Gating.}
With the exception of the input-level concatenation baseline from DDM~\citep{debortoli2025distributional}, every mechanism passes $\xi_j$ through a shared adaptive multiplicative gate
\begin{equation}
    \label{eq:adaptive_ksi_gating}
    \tilde{\xi}_j = \xi_j \odot (W_\text{gate}\, c + 1)\,,
\end{equation}
where $W_\text{gate}$ is a zero-initialized linear layer mapping the conditioning vector $c \in \mathbb{R}^{d_\text{cond}}$ (a combined embedding of timestep $t$ and, in the case of class-conditional image generation, the class) to a per-element multiplicative gate.
Zero initialization ensures that the gate starts at $1$ (identity scaling), so the model initially behaves identically to standard DDM model~\citep{debortoli2025distributional} and gradually learns to modulate $\xi$ as needed.

\subsection{Conditioning Mechanisms}
\paragraph{Channel-wise Concatenation.}
Concatenation appends $\tilde{\xi}_j$ to the residual stream along the channel dimension at layer $\ell_\text{start}$.
In the simplest setting, this results in a widening of the per-token feature width from $d$ to $d + d_\mathrm{cat}$.
This widened residual stream is processed by all subsequent layers, and the velocity predictions are read out \emph{only} from the original channels of the final hidden state as
\begin{equation}
    \hat{v}_\theta \;=\; \mathrm{Unembed}\bigl(h_L[\,\ldots,\,:d\,]\bigr).
\end{equation}
We study three variants of this family:
\begin{itemize}[leftmargin=*]
    \item \textbf{Cat (widen).} All projections in layers $\ell_\text{start},\ldots,L-1$ operate at the widened width $d+d_\mathrm{cat}$. Since attention QKV and FFN intermediate dimensions are typically determined based on the residual stream width, all are enlarged proportionally. This is the most expressive variant, but adds extra learnable parameters and FLOPS in the late layers, conflating results with overall improved capacity.
    \item \textbf{Cat (fixed inner).} The residual stream is widened to $d + d_\mathrm{cat}$ as before, but each block's inner computation is held at the original width. The signal $\tilde{\xi}_j$ can persist in the residual stream across layers, while the bulk of the computation is unchanged. This is our default, as it preserves the approx. parameter count of the baseline.
    \item \textbf{Cat (substitute).} The final variant preserves the number of trainable parameters and the computation of the distributional layers \emph{exactly} identical. This is achieved by \emph{not} widening the residual stream to accommodate $d_\mathrm{cat}$, but instead narrowing the previous residual information to $d - d_\mathrm{cat}$ dimensions (by dropping a set of dimensions) before concatenation.
\end{itemize}

\paragraph{Residual Addition.}
Instead of concatenating the gated noise, it is directly added to the residual stream at layer $\ell_\text{start}$.
To project it to the model width, a linear map $W_\text{add} \in \mathbb{R}^{d_\text{cat} \times d}$ is used, resulting in the residual stream update
\begin{equation}
    h \;\leftarrow\; h + \tilde{\xi}_j W_\text{add}.
\end{equation}
This preserves both width and parameter count of the late layers, with only the linear map introducing additional learnable parameters.

\paragraph{Adaptive Normalization (Ada-Norm).}
$\tilde{\xi}_j$ is mapped through a per-token MLP to scale (and, optionally, shift, if using LayerNorm instead of RMSNorm) parameters that are added to the existing adaptive pre-norms typically used for timestep and class conditioning in diffusion transformers.
This keeps the residual stream width unchanged.

\paragraph{Register Tokens/Sequence-wise Concatenation.}
$\tilde{\xi}_j$ is projected and reshaped into $n_\mathrm{tok}$ additional sequence tokens with learned positional embeddings, and appended to the token stream at layer $\ell_\text{start}$.
The extra tokens participate in self-attention but are dropped from the velocity readout, similar to the additional channels introduced in channel-wise concatenation.

\paragraph{Concatenation at Input (Na\"ive Baseline).}
Finally, we also consider the original DDM formulation~\citep{debortoli2025distributional}, where $\tilde{\xi}_j$ is reshaped to the spatial grid of $x_t$ and concatenated to it along the channel dimension before the patch embedding, with all $L$ transformer layers processing the noise.
This variant is incompatible with deferred expansion ($\ell_\text{start} > 0$) and is reported as a reference only; it is strictly dominated by the post-expansion variants in our experiments.

\paragraph{Discussion.}
\Cref{tab:xi_conditioning_hparam_ablation} shows that fixed-inner concatenation performs substantially better than Ada-Norm and token conditioning in the few-step regime. We hypothesize that, with only a few particle-specific layers after expansion, directly carrying $\xi$ in the residual stream makes the auxiliary signal easier to preserve than routing it indirectly through normalization parameters or attention.

Fixed-inner concatenation gives every spatial token a persistent particle-specific feature subspace throughout the remaining residual blocks, while leaving the expensive internal attention and MLP dimensions unchanged. Additive conditioning instead superposes the auxiliary signal onto the shared representation, making it easier for subsequent transformations to attenuate or entangle that signal. Ada-Norm routes ($\xi$) indirectly through modulation parameters, while token conditioning requires attention to propagate particle information from a small number of auxiliary tokens to every spatial token. This may be difficult when only a few distributional layers remain after expansion.

These mechanisms are all reasonable a priori;~\Cref{tab:xi_conditioning_hparam_ablation} provides the empirical reason for selecting fixed-inner concatenation. We present this explanation as an hypothesis consistent with the observed results, rather than as a theoretically established mechanism.

\section{Kernel Spatial Modes}
\label{app:kernel_modes}

The generalized energy score in \cref{eq:kernel_score} is defined for vector-valued predictions, but in our setting, each prediction $\hat{x}_\theta(t, x_t, \xi_j) \in \mathbb{R}^{H' \times W' \times C}$ is a sequence of patch tokens.
The score, therefore, admits two natural instantiations that differ in the geometry over which the kernel norm $\|\cdot\|_2^\beta$ is evaluated.

\paragraph{Global Kernel.}
Each prediction is flattened into a single $H' W' C$-dimensional vector and the score $S_{\lambda,\beta}$ is evaluated on the resulting vector space:
\begin{equation}
    S_{\lambda,\beta}^\text{global}(p, y) \;=\; -\,\mathbb{E}_{p}\!\left[\|\mathrm{vec}(X) - \mathrm{vec}(y)\|_2^{\beta}\right]
    + \tfrac{\lambda}{2}\,\mathbb{E}_{p\otimes p}\!\left[\|\mathrm{vec}(X) - \mathrm{vec}(X')\|_2^{\beta}\right].
\end{equation}
This is the implicit choice in most MMD/energy-distance work~\citep{Gneiting01032007,debortoli2025distributional} -- it imposes a single global geometry over the entire prediction.

\paragraph{Local Kernel.}
Alternatively, the score is evaluated independently on each spatial position ${X_{h,w} \in \mathbb{R}^{C}}$ and averaged over positions,
\begin{equation}
    S_{\lambda,\beta}^\text{local}(p, y) \;=\; \frac{1}{H' W'} \sum_{h, w}
    S_{\lambda, \beta}\bigl(p_{h,w},\, y_{h,w}\bigr),
\end{equation}
which corresponds to a separable product kernel over the spatial grid.
Each per-position score is strictly proper on its slice of $\mathbb{R}^C$, so the average is also strictly proper as a scoring rule for the per-position marginals of $p$.
The local mode emphasizes the local structure of the velocity prediction, at the cost of decoupling the score from cross-spatial dependencies.
We use the local mode by default since it offers significant practical performance improvements; the comparison is reported in \cref{tab:kernel}.

\section{Training-time $t$ Sampling}
\label{app:t_sampling}
The DDM loss in \cref{eq:ddm_loss} composes a per-time score with a sampling density $p(t)$ and a loss weighting $w(t)$.
We hold $w(t) \equiv 1$ throughout and ablate only the sampling density.
Since reweighting the sampling density without compensating in $w(t)$ amounts to an effective $t$-dependent loss weighting, this also controls the relative supervision allocated to each noise level.

We compare four samplers, three of which belong to the logit-normal family $t = \sigma(\mu + s Z)$ with $Z \sim \mathcal{N}(0,1)$ and $\sigma$ the logistic sigmoid:
\begin{itemize}[leftmargin=*]
    \item \texttt{uniform}: $t \sim \mathcal{U}(0,1)$. Unbiased baseline with constant density on $[0,1]$.
    \item \texttt{logit-normal} ($\mu=0,\,s=1$): the default of \citet{esser2024scaling}; symmetric and peaked at $t=0.5$.
    \item \texttt{jit} ($\mu=-0.8,\,s=0.8$): the sampler used by \citet{li2025jit}; mode at $t \approx 0.25$, shifted toward the noise end (low SNR).
    \item \texttt{imf} ($\mu=-0.4,\,s=1$): the sampler used by \citet{geng2026improved}; mode at $t \approx 0.32$, with a similar bias to \texttt{jit} but broader support.
\end{itemize}
In terms of the anchors of Table~\ref{tab:characteristic-times}, jit places its mode at $t\approx0.25$ ($\rho\approx0.1$), inside Regime~IIIa. It allocates $72.5\%$ of training times to $\rho\in[\rho_c,1]$, compared with $35.2\%$ for uniform and $46.0\%$ for logit-normal$(0,1)$. This is the low-SNR part of the post-collapse regime, where the posterior is mode-dominated but still broad and where our schedule moves the objective from distributional toward regression-like behaviour.~\cref{tab:train_objective_params_ablation}c shows that this allocation improves both few- and many-step FID.

The FM reference and Configs A-E use the \texttt{logit-normal} sampler. Only F switches to \texttt{jit}. The sampler-matched comparison is FM and Config E and therefore isolates the distributional components, which reduce 4-step FID from 26.53 to 14.33 (about 91\% of the total 4-step gain).

\section{MMD lineage}
\label{sec:mmd-lineage}

 The generalized energy score (Eq.~\ref{eq:kernel_score}) can be modified to yield a generalized kernel score (see~\citep{debortoli2025distributional}) by replacing $-||x-y||_{2}^{\beta}$ by a positive definite kernel $k(x,y)$. Such kernel score
then reduces to a Dirac-target MMD objective up to a target-dependent constant when $\lambda{=}1$; 
MMD-based generative training has a long history spanning GMMNs~\citep{li2015generative} and early kernel score objectives such as DISCO Nets~\citep{NIPS2016_c0e190d8}. IMM~\citep{zhou2025inductive} is the population-vs-population MMD generalisation in which the right-hand argument is itself a population, with Consistency Models~\citep{pmlr-v202-song23a, song2024improved} appearing as a single-particle, first-moment limit in the IMM analysis. DDMs and IMM are therefore instances of the same kernel score family with different right-hand arguments; our proposed schedules clarify this connection.

\begin{algorithm}[t]
  \caption{DDM training with deferred population expansion. The first $\ell_\text{start}$ layers run once at batch size $B$; after population expansion and $\xi$ conditioning, the final $H$ layers run at effective batch size $B\times m$.}
  \label{alg:ddm_training}
  \begin{lstlisting}[style=pythonalg]
  # model: DistributionalTransformer with L layers; expansion at layer L-H
  # kernel: GeneralizedKernelScore with schedules lambda(t), beta(t)

  # FM time; t=0 noise, t=1 data
  t  = sample_t(B)                              
  x0 = randn_like(x1)
  xt = (1 - t) * x0 + t * x1
  # one xi per population member
  xi = sample_xi(B * m)                         

  h, cond = model.embed(xt), model.condition(t, y)
  
  # shared trunk at batch B
  for layer in model.layers[:L-H]:              
      h = layer(h, cond)

  # population expansion
  h    = repeat(h,    "b ... -> (b m) ...", m=m)   
  cond = repeat(cond, "b ... -> (b m) ...", m=m)
  # AdaLN / additive / concat / tokens
  h, cond = inject_xi(h, cond, xi)              

  # repeated head at batch B * m
  for layer in model.layers[L-H:]:              
      h = layer(h, cond)
  v_pred = model.unembed(h)

  v_pred = rearrange(v_pred, "(b m) ... -> b m ...", m=m)
  tgt    = repeat(x1 - x0,   "b ... -> b m ...", m=m)
  loss   = -kernel.score(v_pred, tgt, t)
  loss.mean().backward()
\end{lstlisting}
\end{algorithm}

\section{Compute Requirements}
All iDDM models were trained on ImageNet-256$^2$ with 4 $\times$ H200 for B-sized models for roughly 12 hours and 8 $\times$ H200 for XL-sized models. Reproduced models may differ in computational requirements based on their additional training overhead.

\section{Additional Visualizations}
\begin{figure}[h!]
    \centering
    \includegraphics[width=\linewidth]{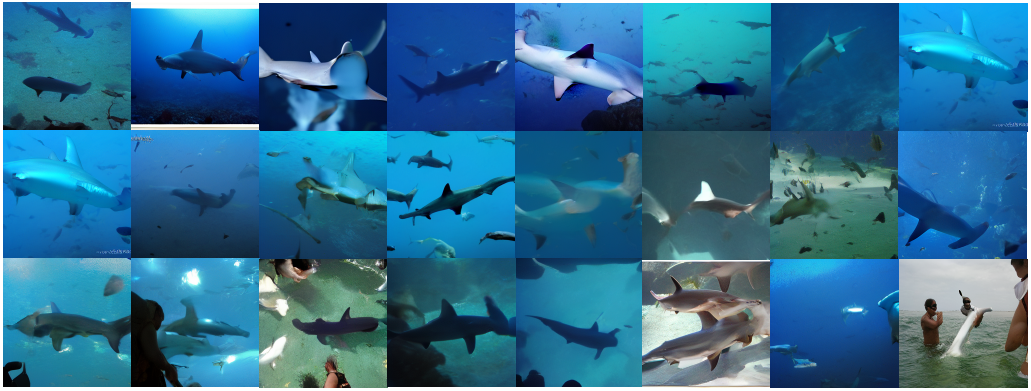}
    \caption{Uncurated ImageNet-256$^2$ samples using 4 steps.}
    \label{fig:uncurated_start}
\end{figure}

\begin{figure}[h!]
    \centering
    \includegraphics[width=\linewidth]{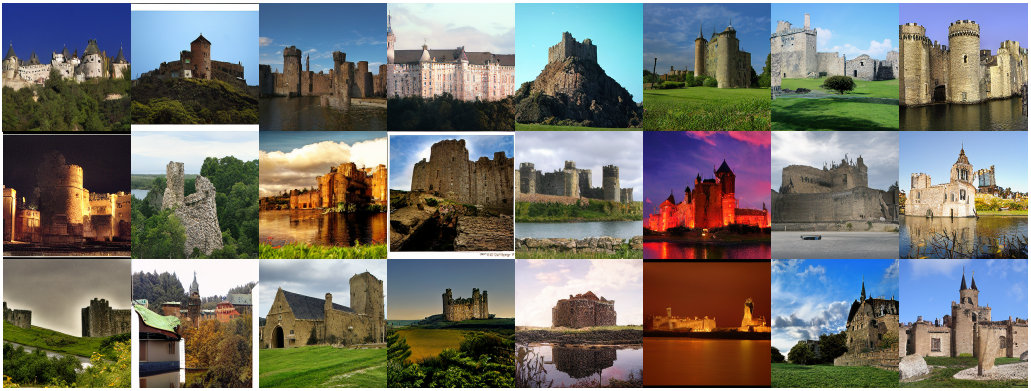}
    \caption{Uncurated ImageNet-256$^2$ samples using 50 steps.}
    \label{fig:uncurated_end}
\end{figure}

We provide additional uncurated samples using 4 and 50 steps, respectively, in \Cref{fig:uncurated_start}-\Cref{fig:uncurated_end}. 
\label{app:visualizations}

\begin{figure}[h!]
    \centering
    \includegraphics[width=\linewidth]{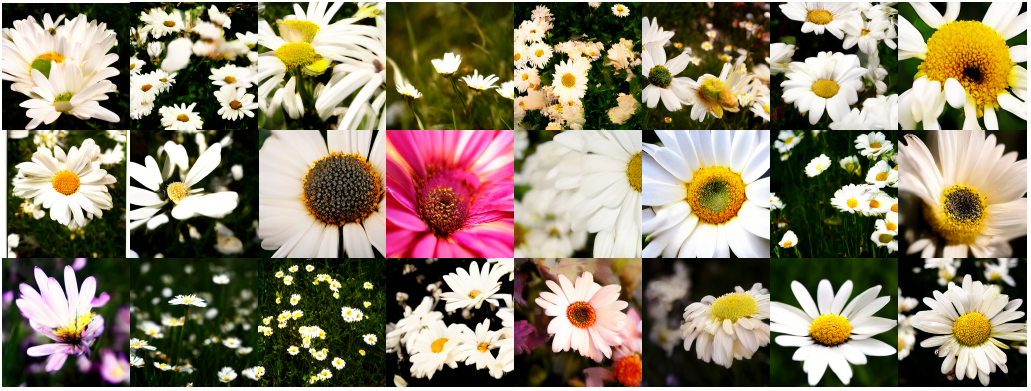}
    \caption{Uncurated ImageNet-256$^2$ samples using 4 steps.}
\end{figure}
\begin{figure}[h!]
    \centering
    \includegraphics[width=\linewidth]{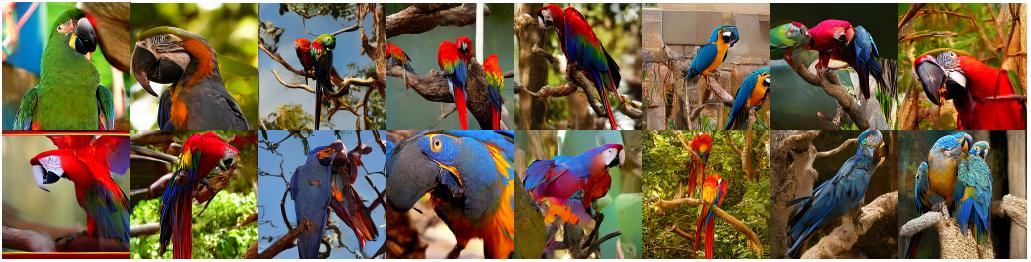}
    \caption{Uncurated ImageNet-256$^2$ samples using 4 steps.}
\end{figure}
\begin{figure}[h!]
    \centering
    \includegraphics[width=\linewidth]{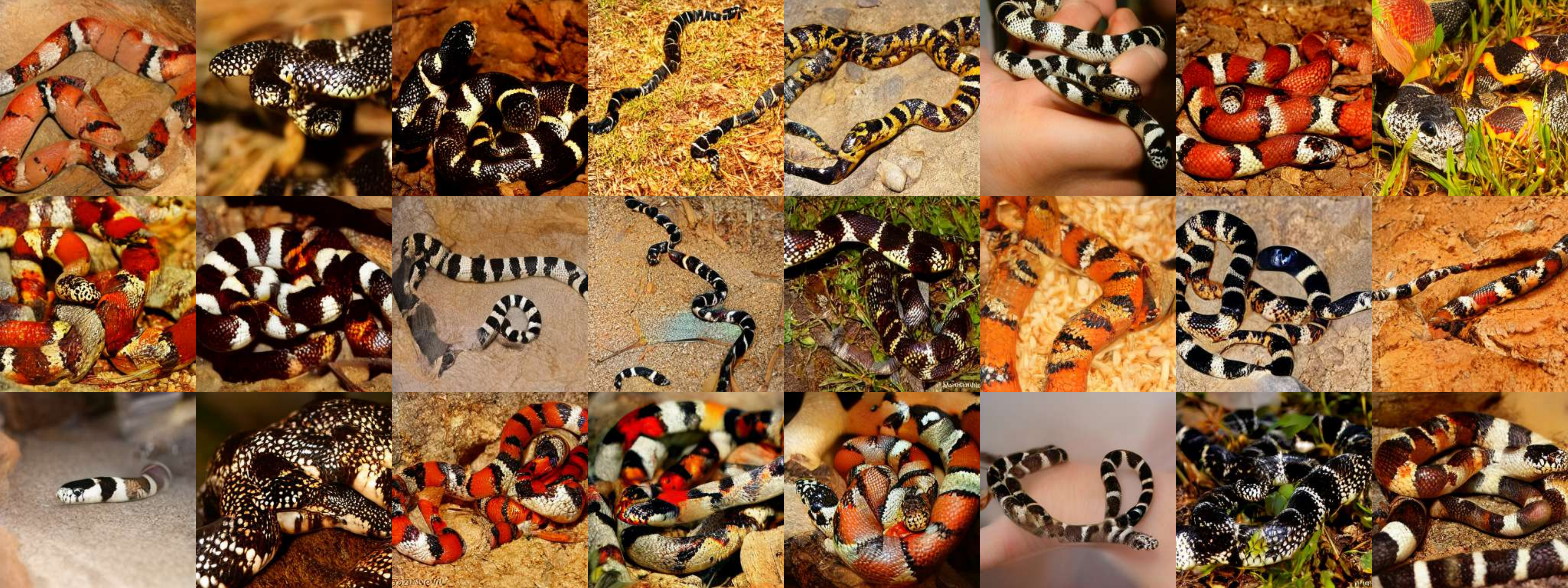}
    \caption{Uncurated ImageNet-256$^2$ samples using 4 steps.}
\end{figure}
\begin{figure}[h!]
    \centering
    \includegraphics[width=\linewidth]{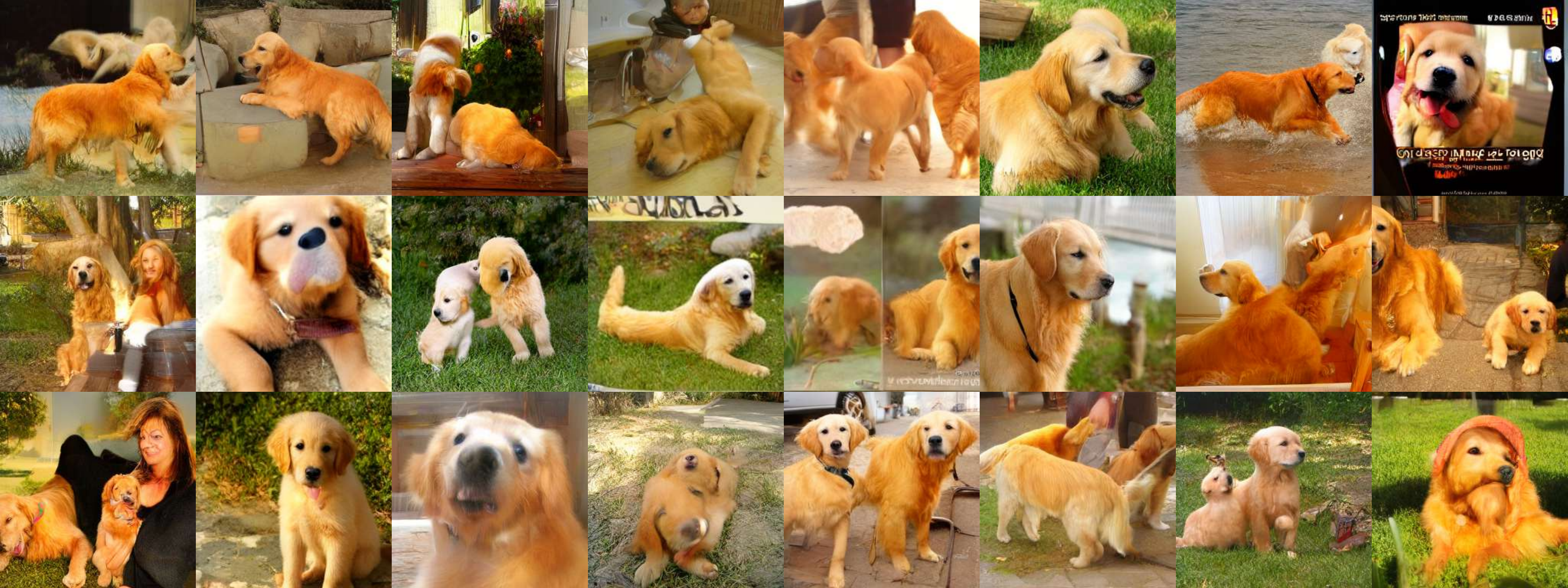}
    \caption{Uncurated ImageNet-256$^2$ samples using 4 steps.}
\end{figure}
\begin{figure}[h!]
    \centering
    \includegraphics[width=\linewidth]{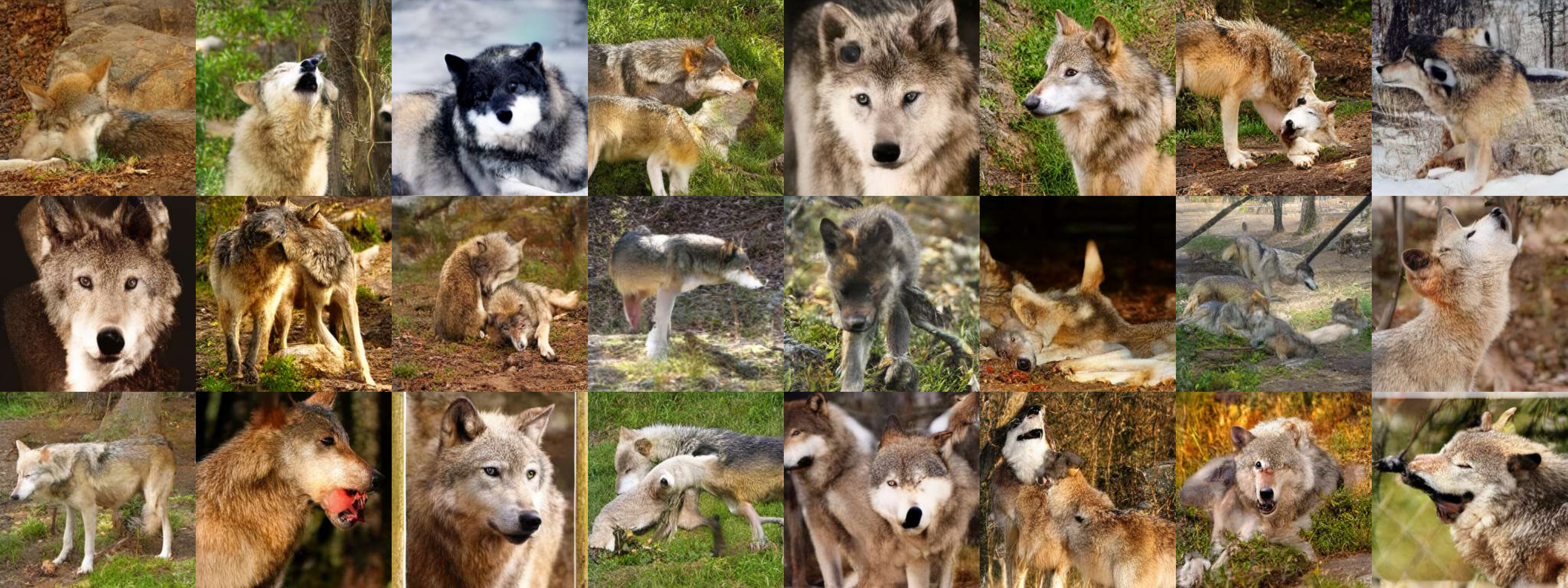}
    \caption{Uncurated ImageNet-256$^2$ samples using 4 steps.}
\end{figure}
\begin{figure}[h!]
    \centering
    \includegraphics[width=\linewidth]{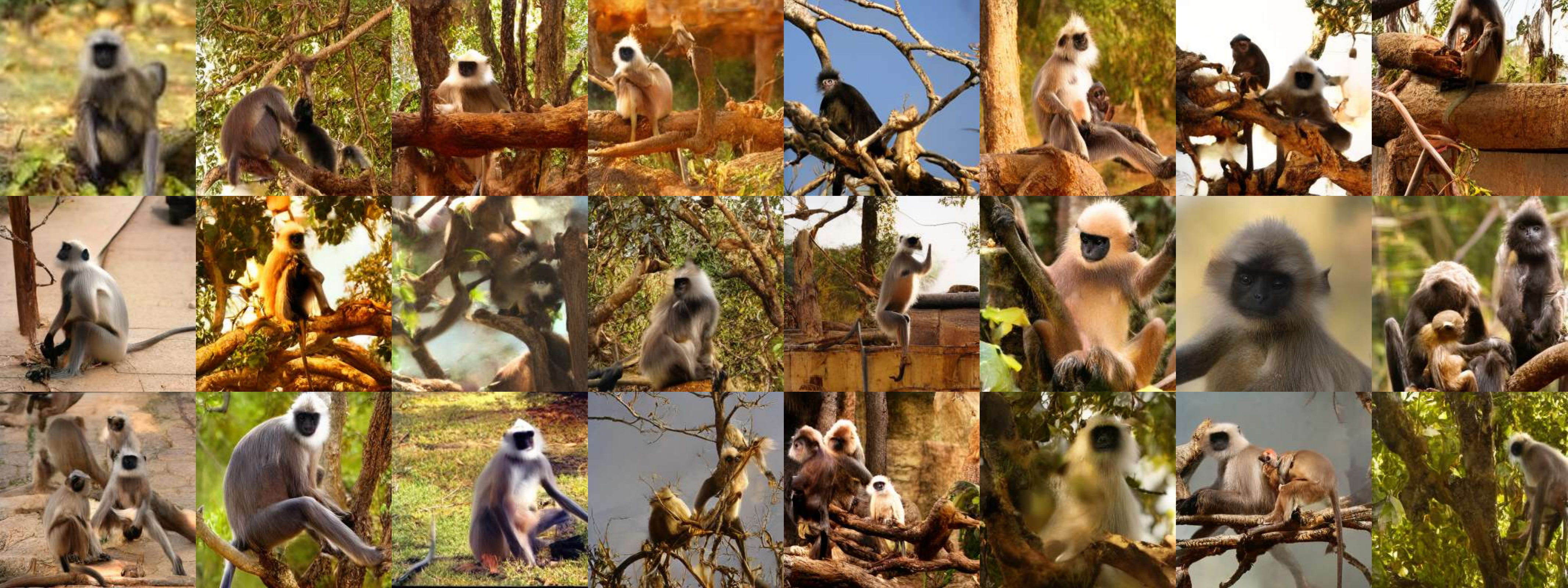}
    \caption{Uncurated ImageNet-256$^2$ samples using 4 steps.}
\end{figure}
\begin{figure}[h!]
    \centering
    \includegraphics[width=\linewidth]{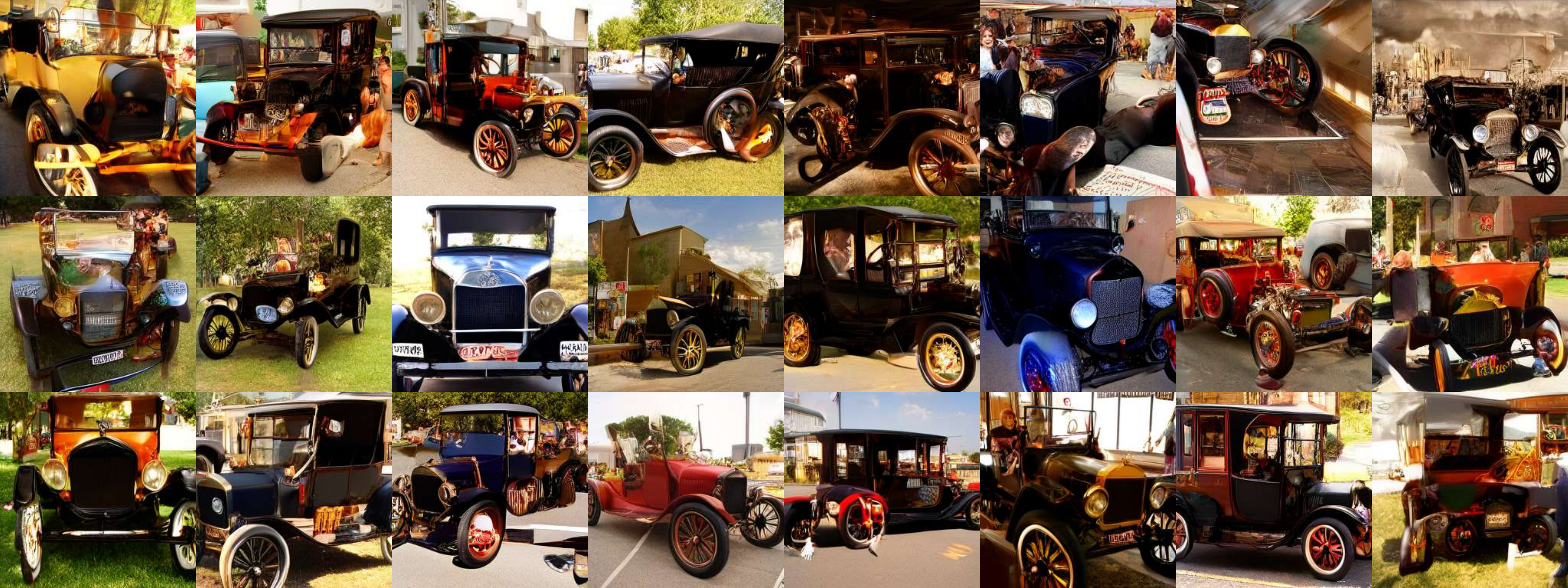}
    \caption{Uncurated ImageNet-256$^2$ samples using 4 steps.}
\end{figure}
\begin{figure}[h!]
    \centering
    \includegraphics[width=\linewidth]{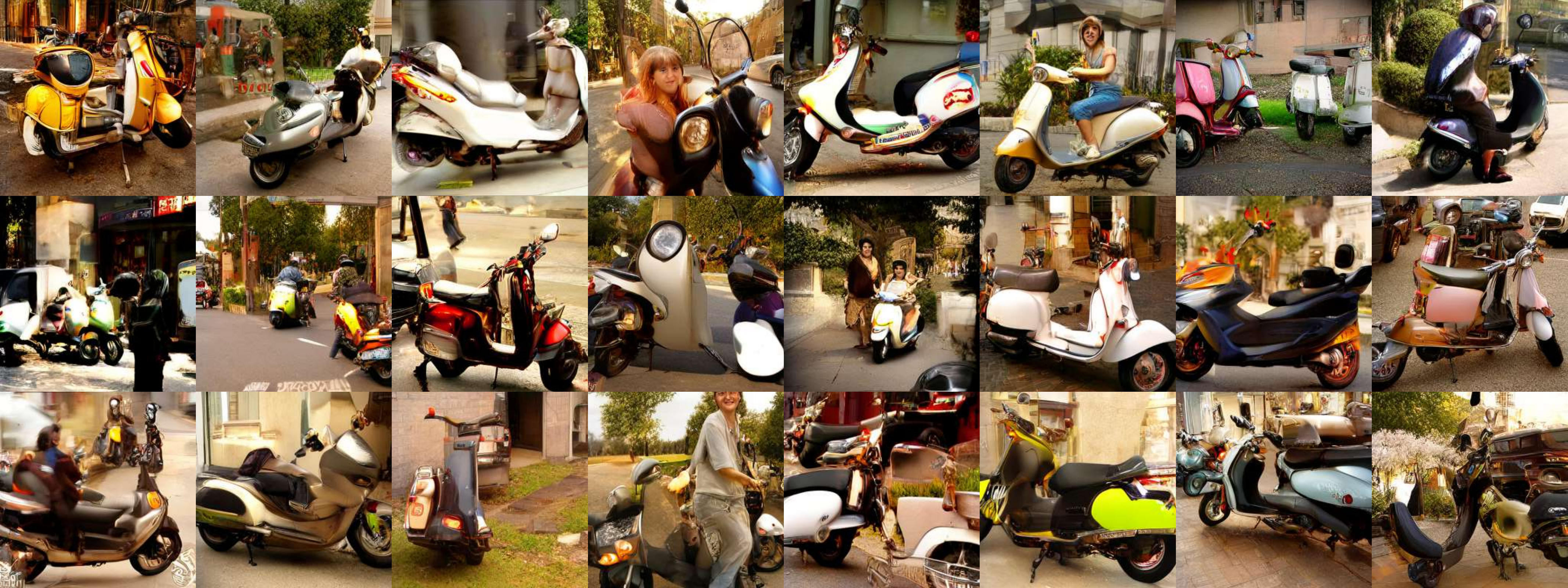}
    \caption{Uncurated ImageNet-256$^2$ samples using 4 steps.}
\end{figure}

\begin{figure}[h!]
    \centering
    \includegraphics[width=\linewidth]{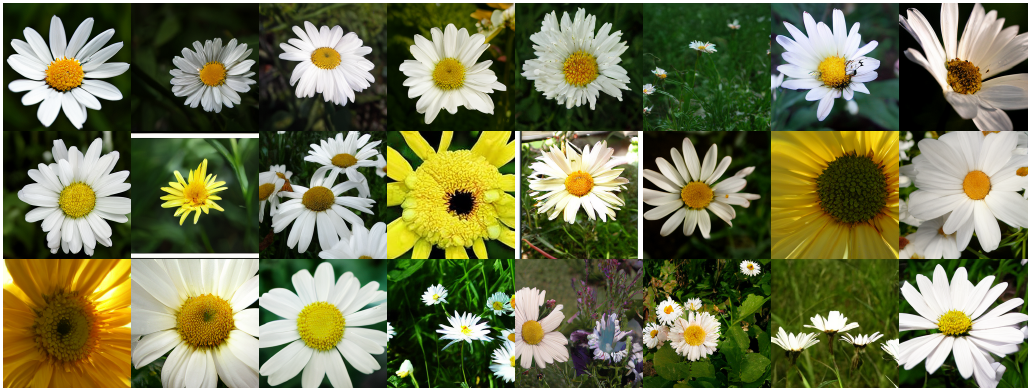}
    \caption{Uncurated ImageNet-256$^2$ samples using 50 steps.}
\end{figure}
\begin{figure}[h!]
    \centering
    \includegraphics[width=\linewidth]{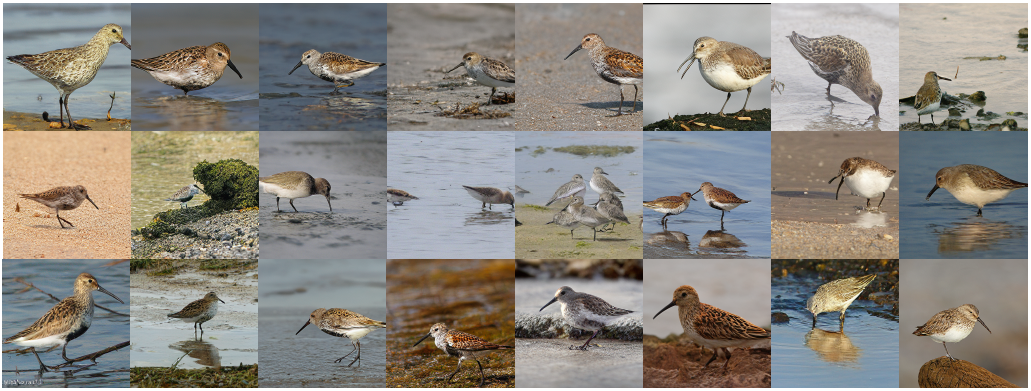}
    \caption{Uncurated ImageNet-256$^2$ samples using 50 steps.}
\end{figure}
\begin{figure}[h!]
    \centering
    \includegraphics[width=\linewidth]{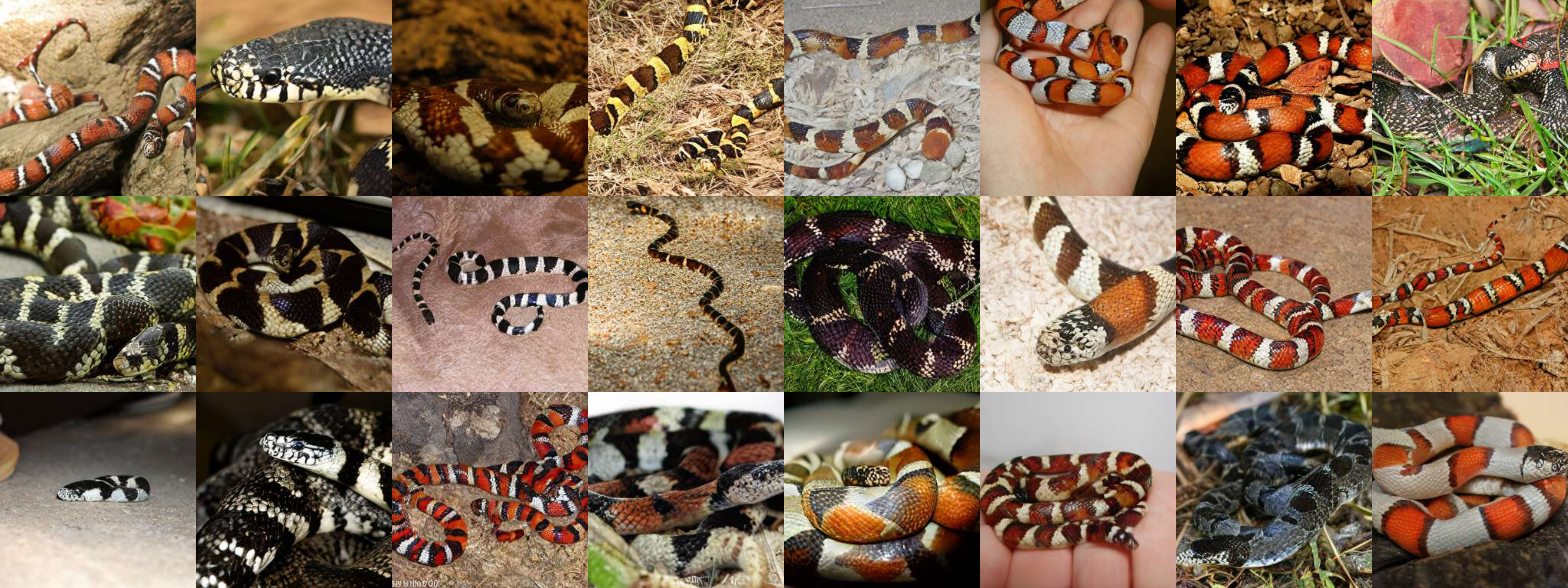}
    \caption{Uncurated ImageNet-256$^2$ samples using 50 steps.}
\end{figure}
\begin{figure}[h!]
    \centering
    \includegraphics[width=\linewidth]{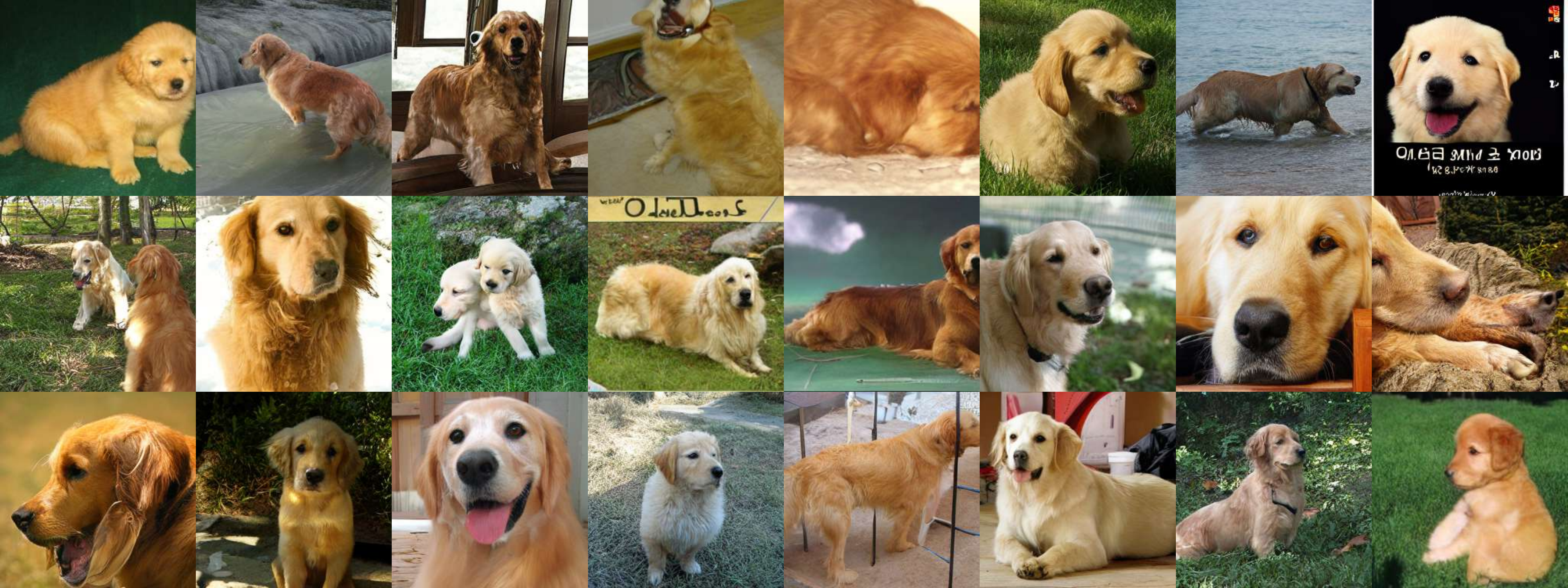}
    \caption{Uncurated ImageNet-256$^2$ samples using 50 steps.}
\end{figure}
\begin{figure}[h!]
    \centering
    \includegraphics[width=\linewidth]{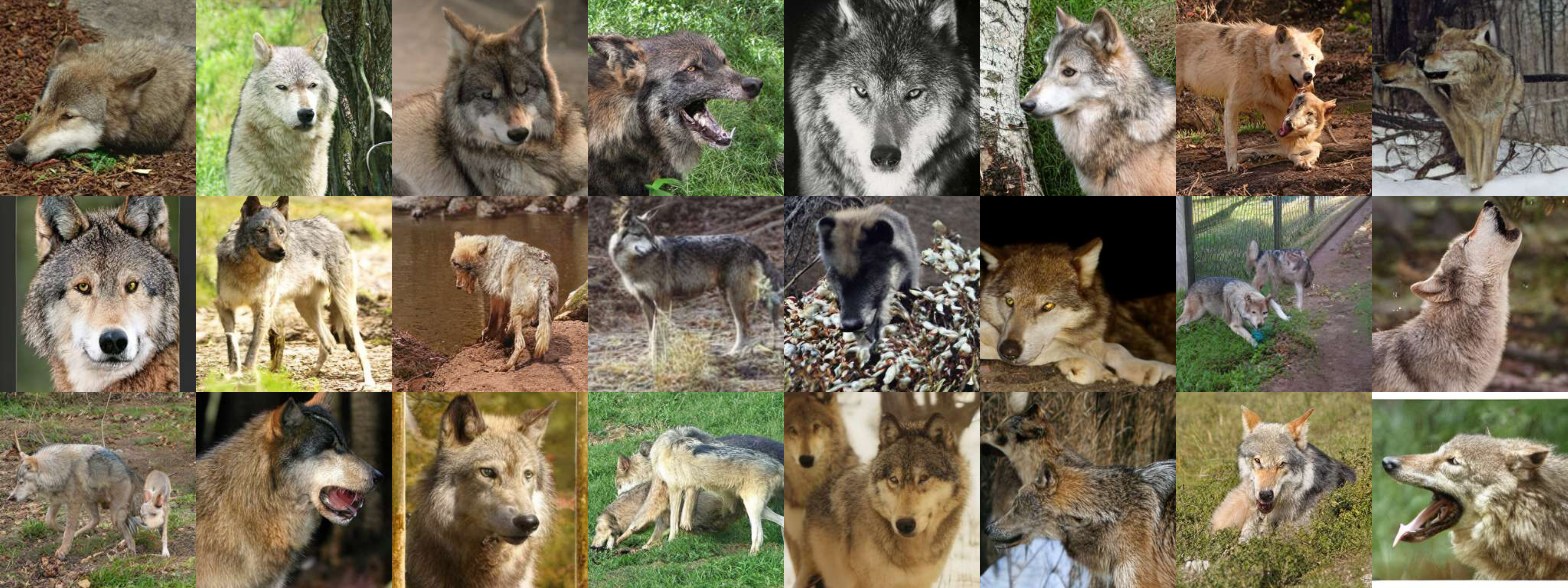}
    \caption{Uncurated ImageNet-256$^2$ samples using 50 steps.}
\end{figure}
\begin{figure}[h!]
    \centering
    \includegraphics[width=\linewidth]{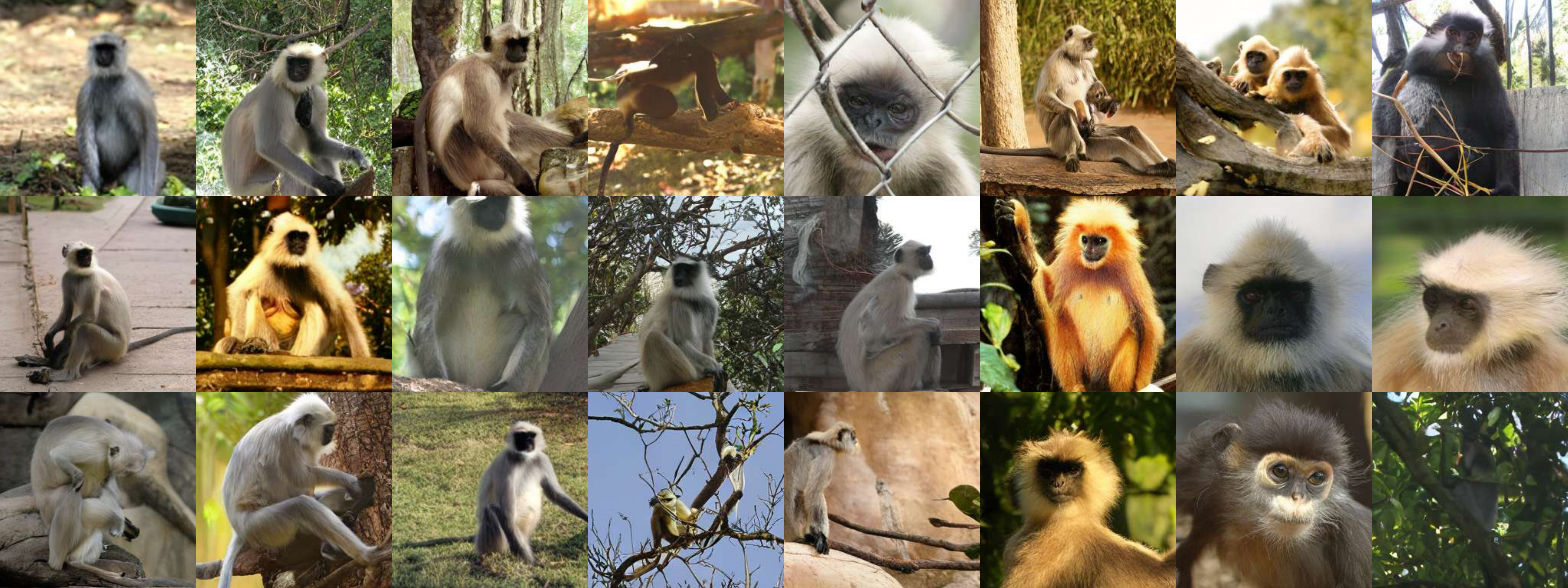}
    \caption{Uncurated ImageNet-256$^2$ samples using 50 steps.}
\end{figure}
\begin{figure}[h!]
    \centering
    \includegraphics[width=\linewidth]{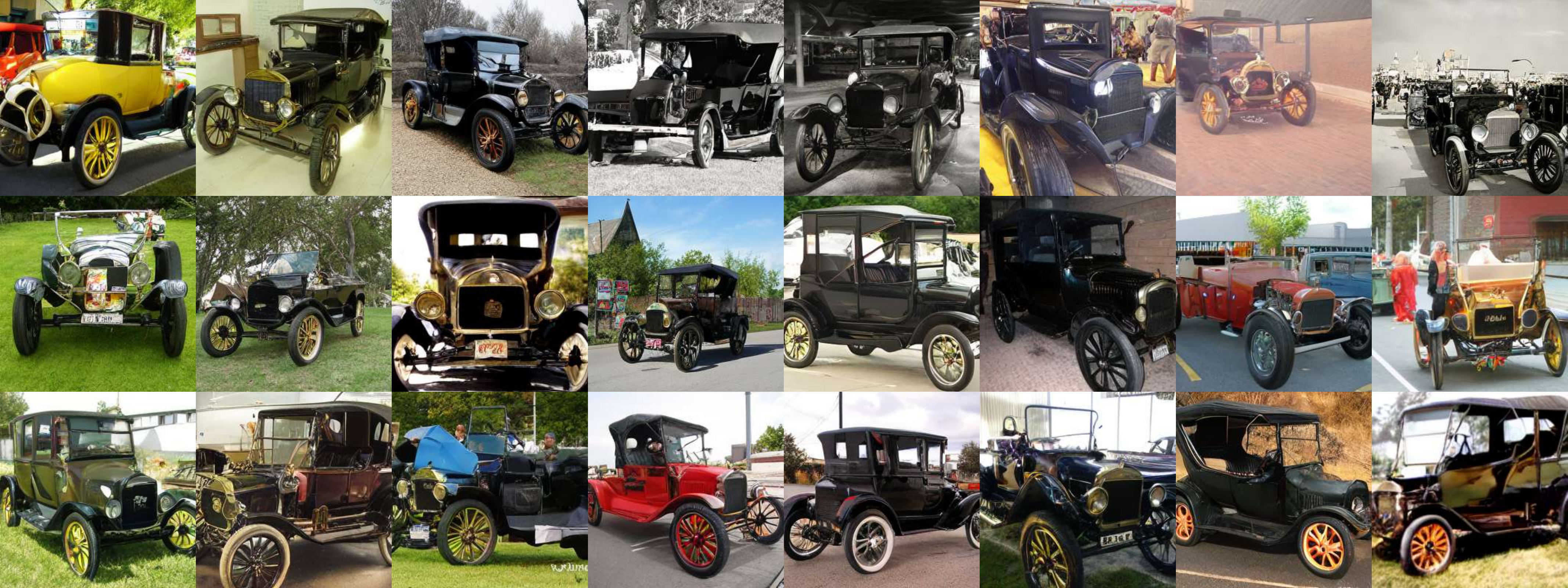}
    \caption{Uncurated ImageNet-256$^2$ samples using 50 steps.}
\end{figure}
\begin{figure}[h!]
    \centering
    \includegraphics[width=\linewidth]{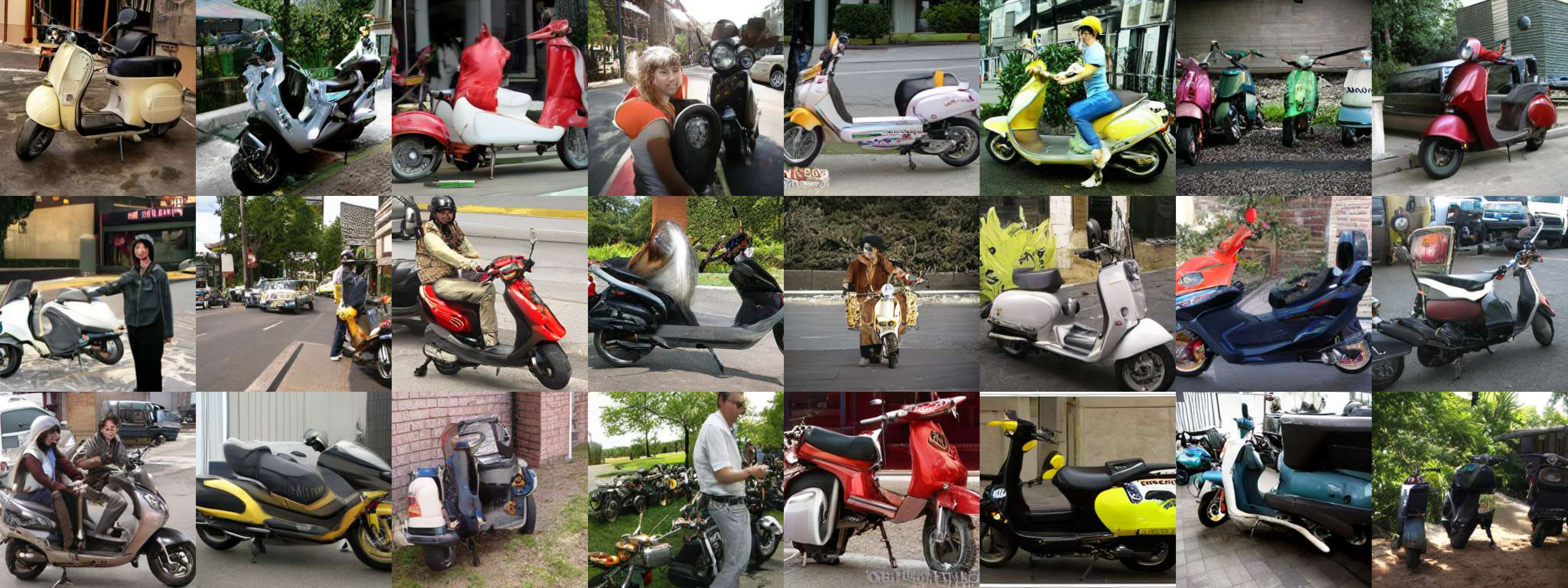}
    \caption{Uncurated ImageNet-256$^2$ samples using 50 steps.}
\end{figure}

\end{document}